\documentclass{article}

\usepackage[preprint]{neurips_2026}

\usepackage{pgfplots}
\pgfplotsset{compat=1.18} 
\usepackage{xcolor}

\definecolor{fedmpo_red}{HTML}{8C1515}   
\definecolor{nocomp_gray}{HTML}{4F6272}  
\definecolor{noalign_blue}{HTML}{B7C3F3} 
\definecolor{nomoe_orange}{HTML}{DD7E6B} 
\definecolor{nomask_green}{HTML}{8FA87A} 

\definecolor{color1}{HTML}{3D7CB4} 
\definecolor{color2}{HTML}{64A071} 
\definecolor{color3}{HTML}{C54576} 
\definecolor{color4}{HTML}{9078B4} 
\definecolor{color5}{HTML}{D15C53} 
\definecolor{colorOurs}{HTML}{8562A2} 

\definecolor{c1D}{HTML}{3D7CB4} \definecolor{c1M}{HTML}{8EBAE5} \definecolor{c1L}{HTML}{D1E5F0} 
\definecolor{c2D}{HTML}{64A071} \definecolor{c2L}{HTML}{A6D0AA} 
\definecolor{c3D}{HTML}{C54576} \definecolor{c3L}{HTML}{DFA6C1} 
\definecolor{c4D}{HTML}{9078B4} \definecolor{c4L}{HTML}{CFC1DF} 
\definecolor{c5D}{HTML}{D15C53} \definecolor{c5L}{HTML}{F6A78D} 
\definecolor{oursD}{HTML}{8562A2} \definecolor{oursM}{HTML}{B9A2C8} \definecolor{oursL}{HTML}{E2D9EB} 

\usepackage[utf8]{inputenc}
\usepackage[T1]{fontenc}
\usepackage{hyperref}
\usepackage{url}
\usepackage{booktabs}
\usepackage{amsfonts}
\usepackage{nicefrac}
\usepackage{microtype}
\usepackage{xcolor}
\usepackage{graphicx}

\usepackage{amsmath}
\usepackage{amssymb}
\usepackage{multirow}
\usepackage{subcaption}
\usepackage{pgfplots}
\usepackage{subcaption}
\usepackage{booktabs}
\usepackage{colortbl} 
\usepackage{tabularx}
\usepackage{DejaVuSans}
\usepackage[scaled]{helvet}
\usepackage{xcolor}
\usepackage{algorithm}
\usepackage{algorithmic}

\floatstyle{plain}
\restylefloat{algorithm}

\usepgfplotslibrary{groupplots}

\title{Towards Robust Federated Multimodal Graph Learning under Modality Heterogeneity}

\author{
Sirui Zhang \quad Haonan Wang \quad Xunkai Li \quad Zekai Chen \quad Shumeng Li 
\\\textbf{Hongchao Qin \quad Rong-Hua Li \quad Guoren Wang} \\
Beijing Institute of Technology 
}

\begin{document}

\maketitle

\begin{abstract}
Recently, multimodal graph learning (MGL) has garnered significant attention for integrating diverse modality information and structured context to support various network applications. However, real-world graphs are often isolated due to data-sharing limitations across multiple parties, and their modalities are frequently incomplete. This highlights an urgent need to develop a robust federated approach. However, we find that existing methods remain insufficient. On the one hand, centralized MGL methods that handle missing modalities overlook the knowledge sharing and generalization in federated scenarios. On the other hand, while federated MGL methods have become increasingly mature, they primarily target non-graph data. Based on these technologies, we identify a two-stage pipeline wherein client-side completion reconstructs missing modalities, and server-side aggregation integrates the client-updated parameters of both the modality generator and the backbone models. Although this serves as a general solution, we identify two primary challenges in achieving greater robustness: \textbf{(1) Topology-Isolated Local Completion}: Client-side modality generation struggles to effectively leverage global semantics. \textbf{(2) Reliability-Imbalanced Global Aggregation}: Server-side multi-party collaboration is hindered by client updates with varying modality availability and recovery reliability. To address these challenges, we propose \textsc{FedMPO}, which utilizes topology-aware cross-modal generation to recover missing features using comprehensive graph context, missing-aware expert routing to locally filter out noisy recovered signals, and reliability-aware aggregation to appropriately down-weight unreliable updates. Extensive experiments on 3 tasks across 6 datasets demonstrate that \textsc{FedMPO} outperforms baselines, achieving performance gains of up to 4.10\% and 5.65\% in high-missing and non-IID settings.
\end{abstract}

\section{Introduction}

Multimodal graph learning models real-world systems by jointly exploiting graph topology and heterogeneous modalities such as text and images~\cite{wei2019mmgcn,tao2020mgat,zhu2025mmgraph}.
In practice, such graphs are often isolated across clients due to privacy constraints, commercial competition, or platform-specific ownership, making centralized training infeasible~\cite{mcmahan2017fedavg,li2026mmopenfgl}. 
Federated learning methods~\cite{mcmahan2017fedavg,li2020fedprox} enable collaboration without sharing raw data, but modalities are frequently incomplete because of heterogeneous collection, storage limitations, privacy restrictions, or user behavior differences. 
Therefore, missing-modality robust learning for federated multimodal graphs becomes a crucial yet underexplored problem in realistic distributed graph applications.

\begin{figure*}[t]
    \centering
    \includegraphics[width=\textwidth]{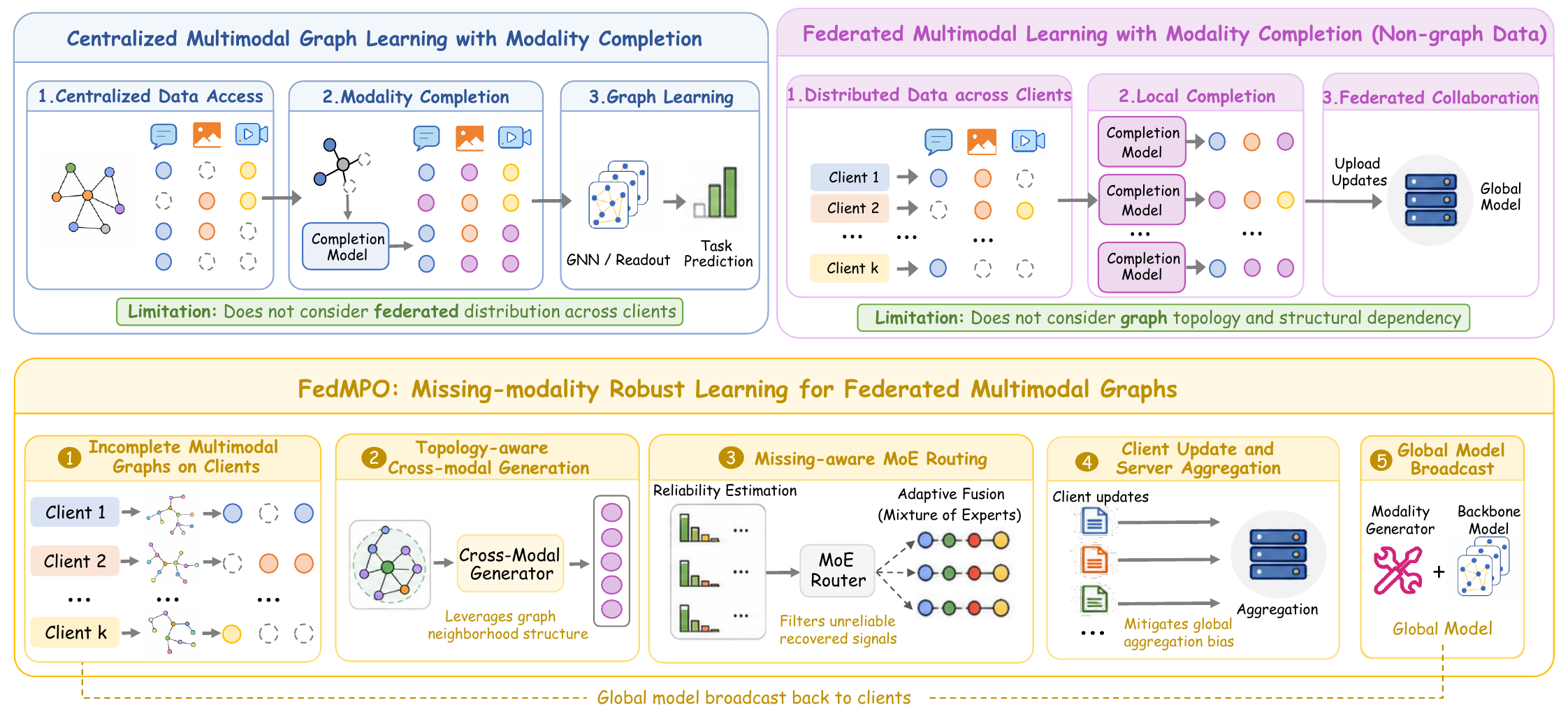}
    \caption{
    \textbf{Overview of missing-modality learning paradigms and the proposed \textsc{FedMPO}.}
    }
    \vspace{-2.0em}
    \label{fig:intro}
\end{figure*}

As shown in Fig.~\ref{fig:intro}, existing studies provide two complementary but incomplete directions. 
Centralized incomplete multimodal graph learning exploits graph topology and multimodal attributes to generate missing modalities before downstream graph learning~\cite{jia2023mhgat,he2025unigraph2,wan2026openmag,wang2023shaspec,wu2024missingmodalitysurvey}. 
It uses graph context as evidence for node-level modality generation, but assumes centralized data access and cannot benefit from cross-client knowledge. 
In contrast, federated missing-modality learning methods aggregate locally updated modality generators on the server and broadcast global knowledge back to clients~\cite{che2024fedmvp,nguyen2024fedmac}, improving local modality generation through collaboration. 
However, they mainly target non-graph data and ignore topology-aware graph context. 
When adapted to federated multimodal graphs, they form a generation--aggregation pipeline, where clients conduct local modality generation and graph learning while the server aggregates modality-generator and graph-backbone updates. 
Since each client only observes its own subgraph, missing features are generated from locally biased graph context and partial modality observations rather than globally consistent cross-client structural semantics. 
Thus, this pipeline remains limited by client-isolated local generation and reliability-skewed global aggregation.

We identify two key challenges in this pipeline. 
\textbf{(1) Topology-Isolated Local Completion}, where modality generation is performed within client-owned subgraphs and struggles to leverage global semantics beyond local graph context. 
Although local topology provides useful graph context, sparse observations and biased client subgraphs may still generate features that are inconsistent with cross-client structural semantics, further misleading graph message passing. 
\textbf{(2) Reliability-Imbalanced Global Aggregation}. 
Since clients differ in modality availability, missingness patterns, and generation reliability, data-size-based aggregation may bias the shared modality generator and graph backbone toward unreliable generated signals or dominant modality patterns.

To address these challenges, we propose \textsc{FedMPO}, a unified missing-modality robust framework for federated multimodal graphs. 
\textsc{FedMPO} follows a four-stage pipeline. 
\textbf{(1) Client-side feature and graph context encoding} distinguishes natural missingness from training-time masking and builds neighborhood-guided structural anchors as topology priors for modality generation (Sec.~\ref{subsec:feature_graph_encoding}). 
\textbf{(2) Client-side topology-aware cross-modal generation} recovers missing features by querying cross-modal context banks and graph context, addressing topology-isolated local completion (Sec.~\ref{subsec:topology_completion}). 
\textbf{(3) Client-side missing-aware mixture-of-experts routing} estimates generation uncertainty and fuses observed, generated, and structural signals to filter noisy recovered signals before local update (Sec.~\ref{subsec:reliability_fusion}). 
\textbf{(4) Server-side reliability-aware aggregation} aggregates client-updated modality-generator and graph-backbone parameters, and uses reliability statistics to down-weight unreliable updates, mitigating reliability-imbalanced global aggregation (Sec.~\ref{subsec:federated_optimization}).

\textbf{Our contributions.} 
(1) \textit{\underline{New Problem}}: We study missing-modality robust learning for federated multimodal graphs, a practical yet underexplored problem where graph-structured multimodal data are distributed and often incomplete.
(2) \textit{\underline{New Method}}: We propose \textsc{FedMPO}, which integrates topology-aware cross-modal generation, missing-aware expert routing, and reliability-aware aggregation into a unified client--server workflow to improve local modality generation reliability and global robustness. 
(3) \textit{\underline{SOTA Performance}}: Experiments on three downstream tasks across six datasets under challenging missing and non-IID conditions demonstrate the effectiveness of \textsc{FedMPO}, achieving up to 4.10\% and 5.65\% gains under high-missing and highly non-IID settings, respectively. 

\section{Preliminaries \& Related Work}
\label{sec:preliminaries}

\subsection{Notations and Problem Formulation}
\label{subsec:problem_formulation}

We study missing-modality robust learning on federated multimodal graphs. 
Consider a federated system with $K$ clients and a central server. 
Each client $k$ owns a private multimodal graph $\mathcal{G}_k=(\mathcal{V}_k,\mathcal{E}_k)$, where nodes denote local entities and edges encode relational dependencies. 
Each node $v_{k,i}\in\mathcal{V}_k$ is associated with $M$ modality-specific attributes $\mathbf{x}_{k,i}=\{\mathbf{x}_{k,i}^{(1)},\ldots,\mathbf{x}_{k,i}^{(M)}\}$, where $\mathbf{x}_{k,i}^{(m)}$ denotes the feature of modality $m$. 
Due to privacy constraints and data ownership, clients cannot directly exchange raw node attributes, modality contents, or local graph structures. 
They may also exhibit heterogeneous feature, label, topology, and missingness distributions, leading to coupled non-IID challenges across multimodal graph clients.

To describe incomplete observations, we define a modality-availability mask $\mathbf{r}_{k,i}\in\{0,1\}^{M}$, where $r_{k,i}^{(m)}=1$ indicates that modality $m$ is observed and $r_{k,i}^{(m)}=0$ otherwise. 
Thus, the accessible input of node $v_{k,i}$ consists of its observed modalities and local graph context, denoted as $\widetilde{\mathbf{x}}_{k,i}=\{\mathbf{x}_{k,i}^{(m)} \mid r_{k,i}^{(m)}=1\}$. 
Missingness may occur at both the client level and the node level. 
We distinguish natural missingness from training-time masking: naturally missing modalities have no ground-truth features, while artificially masked observed modalities provide reconstruction supervision.

For a downstream task $\mathcal{T}$, such as node classification, link prediction, or modality retrieval, the learning process follows three steps. 
\textbf{First}, each client performs local learning on its private incomplete multimodal graph by exploiting available modalities and graph neighborhoods. 
\textbf{Second}, missing modalities are recovered or represented locally, but their reliability can vary across clients because of different modality availability and structural context. 
\textbf{Third}, the server aggregates client updates to obtain a global model and broadcasts it back for the next communication round.

This formulation highlights two core requirements. 
Locally, missing-modality recovery should be topology-aware, since node semantics are shaped by both its own observed modalities and its neighbors through local structural context. 
Globally, federated aggregation should be reliability-aware, since data-size-based aggregation may over-emphasize client updates dominated by unreliable recovered signals. 
Therefore, the goal of \textsc{FedMPO} is to learn robust local representations from incomplete multimodal graphs and integrate the resulting client updates into a reliable global model without sharing raw multimodal attributes or graph structures. 
\vspace{-0.2em}

\subsection{Related Work}
\label{subsec:related_works}
\vspace{-0.2em}

\textbf{Graph and multimodal graph learning.}
GNNs learn node representations by aggregating neighborhood information, with representative models including GCN, GraphSAGE, GAT, and GIN~\cite{kipf2017gcn,hamilton2017graphsage,velickovic2018gat,xu2019gin}. 
Multimodal graph learning further incorporates heterogeneous attributes such as text and images into graph representation learning~\cite{wei2019mmgcn,tao2020mgat,jia2023mhgat,he2025unigraph2}. 
In such graphs, topology not only connects related entities but also provides contextual evidence for resolving modality ambiguity and enhancing incomplete node semantics. 
Recent benchmarks highlight the importance of cross-modal alignment, modality fusion, and graph topology in multimodal graph scenarios~\cite{zhu2025mmgraph,wan2026openmag}. 
However, most existing methods assume centralized data access and complete or directly usable modalities, making them unsuitable for privacy-constrained federated settings with missing modalities.

\textbf{Federated graph and multimodal federated learning.}
Federated learning enables collaborative training across distributed clients without raw data sharing, and methods such as FedAvg, FedProx, and SCAFFOLD address optimization under data heterogeneity~\cite{mcmahan2017fedavg,li2020fedprox,karimireddy2020scaffold}. 
Federated graph learning extends this paradigm to graph data, mainly focusing on structural heterogeneity, client drift, and communication efficiency~\cite{li2025openfgl}. 
Multimodal federated learning studies modality heterogeneity and partial-modality clients~\cite{che2024fedmvp,nguyen2024fedmac,peng2024fedmm,wu2024fiarse,xie2024mhpflid}. 
Nevertheless, most FGL methods assume complete node features, while multimodal FL methods usually treat samples independently and ignore graph-structured dependencies. 
Recent benchmark studies further show that naive combinations of FGL and multimodal FL are insufficient for practical multimodal graph scenarios with incomplete modalities~\cite{li2026mmopenfgl}.

\textbf{Missing-modality learning.}
Incomplete multimodal learning handles partially missing modalities through cross-modal reconstruction, generative imputation, shared--specific representation learning, and missing-aware fusion~\cite{ma2021smil,ma2022missingtransformer,wang2023shaspec,reza2024missingpeft,wu2024missingmodalitysurvey}. 
Although effective in centralized or non-graph settings, these methods do not explicitly model neighborhood-dependent semantics or cross-client aggregation bias, especially under heterogeneous client-level modality missingness. 
In multimodal graphs, unreliable recovered features may be propagated through message passing and further amplified by federated aggregation. 
\textsc{FedMPO} addresses this gap by integrating topology-aware completion, missing-aware routing, and reliability-aware aggregation into a unified federated multimodal graph learning framework.
\section{Methodology}
\label{subsec:methodology}
We present \textsc{FedMPO}, a missing-modality robust framework for federated multimodal graph learning. 
Modality incompleteness affects both local graph representation learning and global federated optimization: locally, missing modalities should exploit neighborhood structural and semantic evidence rather than be recovered independently; globally, clients with severe missingness or unreliable recovery may bias model aggregation.

\begin{figure}[t]
    \centering
    \includegraphics[width=\linewidth]{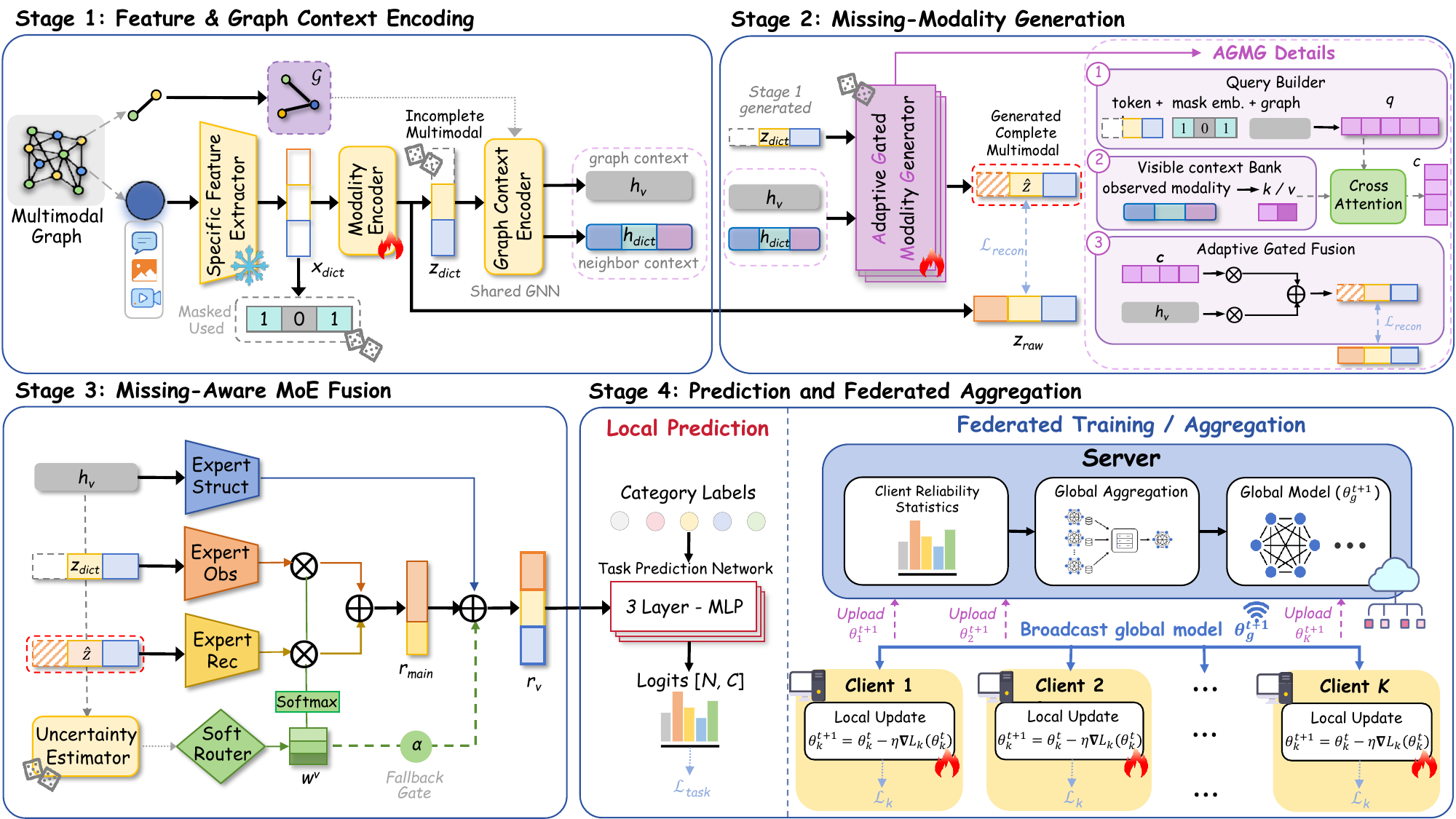}
    \caption{The overview of the proposed \textsc{FedMPO}.}
    \label{fig:fedmpo_framework}
\end{figure}

As shown in Fig.~\ref{fig:fedmpo_framework}, \textsc{FedMPO} uses a four-stage pipeline. 
Each client first encodes available modalities and graph context into a shared space, then generates missing modalities with cross-modal evidence and structural anchors. 
The observed, recovered, and structural signals are further fused by a missing-aware MoE module, followed by local prediction and reliability-aware server aggregation.

\vspace{-0.7em}

\subsection{Stage 1: Feature and Graph Context Encoding}
\label{subsec:feature_graph_encoding}

Each client owns an incomplete multimodal graph, where different nodes may have different available modalities. 
For node $v_i$ and modality $m$, we denote the natural modality availability by $r_i^{(m)}\in\{0,1\}$. 
Since naturally missing modalities have no ground-truth features, they cannot be directly used as reconstruction targets. 
To obtain reliable supervision for modality generation, we apply an artificial mask only to naturally observed modalities. 
The effective visibility mask, artificial reconstruction indicator, and raw modality embedding are defined as
\begin{equation}
\label{eq:mask_raw}
\tilde r_i^{(m)}=r_i^{(m)}\bar r_i^{(m)}, 
\qquad
\Delta_i^{(m)}=r_i^{(m)}-\tilde r_i^{(m)}, 
\qquad
\mathbf z_{i,\mathrm{raw}}^{(m)}=\phi_m(\mathbf x_i^{(m)}).
\end{equation}
Here, $\tilde r_i^{(m)}$ indicates whether modality $m$ is visible to the model, and $\Delta_i^{(m)}=1$ means that an originally observed modality is deliberately hidden for training the generator. 
This design separates natural missingness from training-time masking, so that the model learns from reliable reconstruction targets without assuming unavailable modalities are known.

A key difference between multimodal graphs and ordinary multimodal samples is that node semantics are shaped not only by their own attributes, but also by their neighbors. 
Therefore, before generating missing modalities, \textsc{FedMPO} first constructs a modality-wise structural anchor from neighboring nodes whose modality-$m$ features are visible:
\begin{equation}
\label{eq:anchor}
\mathbf z_{i,\mathrm{anc}}^{(m)}
=
\frac{
\sum_{v_j\in\mathcal N(i)}
a_{ij}\tilde r_j^{(m)}
\mathbf z_{j,\mathrm{raw}}^{(m)}
}{
\sum_{v_j\in\mathcal N(i)}
a_{ij}\tilde r_j^{(m)}
+\epsilon
}.
\end{equation}
Here, $a_{ij}$ denotes normalized neighbor weights, implemented with degree- or entropy-aware normalization to suppress high-degree or redundant neighbors.
The anchor acts as a topology-derived prior for the missing modality. 
If no valid neighbor is available, we use a learnable null token or zero vector as the fallback. 
In this way, a missing modality is not inferred solely from the current node's other modalities, but also from structurally related nodes.

We then feed either the visible raw embedding or its structural anchor into a graph encoder with lightweight modality adapters:
\begin{equation}
\label{eq:graph_context}
\mathbf h_i^{(m)}
=
\mathrm{GNN}_{\theta}
\left(
\psi_m
\left(
\tilde r_i^{(m)}\mathbf z_{i,\mathrm{raw}}^{(m)}
+
(1-\tilde r_i^{(m)})\mathbf z_{i,\mathrm{anc}}^{(m)}
\right),
\mathcal G
\right).
\end{equation}
The output $\mathbf h_i^{(m)}$ is a graph-enhanced representation for modality $m$. 
For reconstructing a target modality, we further pool the representations of other modalities into a target-exclusive context $\mathbf h_{i,\setminus m}$. 
This prevents the generator from trivially copying the target modality during artificial masking and forces it to exploit complementary multimodal and structural evidence.

\subsection{Stage 2: Topology-aware Cross-modal Generation}
\label{subsec:topology_completion}

The second stage aims to recover missing modality representations. 
A naive generator usually predicts a missing modality from the remaining modalities of the same node. 
However, this is insufficient for multimodal graphs because missing semantics may be better inferred from both its own observed modalities and neighboring observed modalities. 
Therefore, \textsc{FedMPO} builds a context bank $\mathcal B_i^{(m)}$ for each target modality $m$, which contains the current node's other visible modalities and neighboring visible modality representations. 
To avoid self-leakage during reconstruction, the target node's own modality-$m$ token is excluded by a diagonal attention mask $\mathbf D_i^{(m)}$, whose corresponding logit is set to $-\infty$ before softmax.

To query this context bank, we construct a modality-specific query using the target-exclusive context, node-level missingness pattern, and target modality identity:
\begin{equation}
\label{eq:query_attention}
\mathbf Q_i^{(m)}
=
\mathbf W_Q
[
\mathbf h_{i,\setminus m}
\parallel
\mathbf E_{\mathrm{mask}}(\tilde{\mathbf r}_i)
\parallel
\mathbf E_{\mathrm{mod}}(m)
],
\qquad
\mathbf c_i^{(m)}
=
\mathrm{MHA}
(
\mathbf Q_i^{(m)},\mathcal B_i^{(m)};\mathbf D_i^{(m)}
).
\end{equation}
The attended evidence $\mathbf c_i^{(m)}$ summarizes useful observed modalities and neighboring signals for recovering the target modality, making the generator topology-aware rather than purely cross-modal.

The generated representation combines attention-based cross-modal evidence, pooled self context, and the structural anchor:
\begin{equation}
\label{eq:generation}
\hat{\mathbf z}_i^{(m)}
=
\gamma_t
\left(
\mathbf g_i^{(m)}\odot \mathbf c_i^{(m)}
+
(\mathbf 1-\mathbf g_i^{(m)})\odot \mathbf W_s\mathbf h_{i,\setminus m}
\right)
+
(1-\gamma_t)\mathbf W_a\mathbf z_{i,\mathrm{anc}}^{(m)} .
\end{equation}
Here, $\mathbf g_i^{(m)}$ is a learnable gate balancing cross-modal evidence and self context. 
We use a linear warmup $\gamma_t=\min(1,t/T_w)$, where $T_w$ is the warmup round. 
Early rounds therefore rely more on structural anchors, which act as a low-variance topology prior and reduce unstable gradients in the non-convex generator optimization under noisy recovered signals. 
As training proceeds, \textsc{FedMPO} gradually shifts toward learned cross-modal generation.

The generation module is supervised only on artificially masked modalities:
\begin{equation}
\label{eq:rec}
\mathcal L_{\mathrm{rec}}
=
\frac{
\sum_{i,m}
\Delta_i^{(m)}
\left\|
\hat{\mathbf z}_i^{(m)}
-
\mathrm{sg}(\mathbf z_{i,\mathrm{raw}}^{(m)})
\right\|_2^2
}{
\sum_{i,m}\Delta_i^{(m)}+\epsilon
}.
\end{equation}
Naturally missing modalities are still generated, but they are not used as direct reconstruction targets because their ground-truth features are unavailable. 
This avoids false supervision in the completion process.

In addition to reconstruction, recovered modalities should be semantically consistent with observed modalities. 
We therefore apply a cross-modal alignment objective:
\small{
\begin{equation}
\label{eq:align}
\mathcal L_{\mathrm{align}}
=
\frac{1}{|\Omega|}
\sum_{(i,m,m')\in\Omega}
\left[
1-
\cos
\left(
\mathbf W_{\mathrm{al}}\tilde{\mathbf z}_i^{(m)},
\mathbf W_{\mathrm{al}}\tilde{\mathbf z}_i^{(m')}
\right)
\right],
\quad
\tilde{\mathbf z}_i^{(m)}
=
\tilde r_i^{(m)}\mathbf z_{i,\mathrm{raw}}^{(m)}
+
(1-\tilde r_i^{(m)})\hat{\mathbf z}_i^{(m)} .
\end{equation}
}
Here, $\Omega$ denotes valid modality pairs. 
This objective encourages observed and recovered modalities of the same node to lie in a coherent semantic space, reducing inconsistency before downstream fusion.

\subsection{Stage 3: Missing-aware Mixture-of-Experts Fusion}
\label{subsec:reliability_fusion}

Although Stage 2 provides candidate representations for missing modalities, not all recovered features are equally reliable. 
For example, a missing modality can be recovered with high confidence when the node has rich observed modalities and many informative neighbors. 
In contrast, recovery may be unreliable when both node-level observations and neighborhood evidence are sparse. 
Directly fusing observed and recovered features without distinguishing their reliability may propagate noisy semantics through graph message passing.

To address this issue, \textsc{FedMPO} first estimates a modality-wise recovery uncertainty and then uses it to control expert routing in a reliability-aware manner:
\begin{equation}
\label{eq:uncertainty_routing}
u_i^{(m)}
=
\begin{cases}
0, & \tilde r_i^{(m)}=1,\\
\sigma
\left(
f_{\theta_u}
[
\hat{\mathbf z}_i^{(m)}
\parallel
\mathbf h_{i,\setminus m}
\parallel
\mathbf z_{i,\mathrm{anc}}^{(m)}
]
\right), & \tilde r_i^{(m)}=0,
\end{cases}
\end{equation}
\begin{equation}
\qquad
[w_{i,\mathrm{obs}}^{(m)},w_{i,\mathrm{rec}}^{(m)}]
=
\mathrm{Softmax}
\left(
f_{\theta_r}
[
\tilde r_i^{(m)}
\parallel
u_i^{(m)}
\parallel
\rho_i
\parallel
\rho
]/\tau
\right).
\end{equation}

Observed modalities are assigned zero recovery uncertainty, while generated modalities are assigned uncertainty scores based on generated features, cross-modal context, and structural anchors. 
The uncertainty head is calibrated on artificially masked observed modalities by matching uncertainty to normalized reconstruction errors. 
For naturally missing modalities without error labels, the calibrated proxy is transferred through shared generator features, while missing-ratio cues reduce heterogeneity-induced shift. 
The router further considers node-level missing ratio $\rho_i$ and client-level missing ratio $\rho$, so that routing decisions are aware of both local and client-wise missingness patterns.

The routed modality representation is produced by two experts: an observed expert and a recovered expert. 
The observed expert focuses on reliable raw modality semantics, while the recovered expert processes generated or recovery-related semantics:
\begin{equation}
\label{eq:expert_representation}
\mathbf f_i^{(m)}
=
\begin{cases}
w_{i,\mathrm{obs}}^{(m)}E_{\mathrm{obs}}(\mathbf z_{i,\mathrm{raw}}^{(m)})
+
w_{i,\mathrm{rec}}^{(m)}E_{\mathrm{rec}}(\hat{\mathbf z}_i^{(m)}),
& \tilde r_i^{(m)}=1,\\
E_{\mathrm{rec}}(\hat{\mathbf z}_i^{(m)}),
& \tilde r_i^{(m)}=0.
\end{cases}
\end{equation}
This design avoids treating observed and recovered modalities as equally trustworthy. 
For observed modalities, the model can still exploit the recovered branch as complementary regularization; for missing modalities, it relies on the recovered expert. 
To avoid routing collapse, we include a load-balancing regularizer in the routing loss, together with uncertainty calibration.

After expert routing, modalities are fused according to their reliability. 
More reliable modalities receive larger weights, while uncertain recovered modalities contribute less. 
For highly incomplete nodes, a structural fallback expert is introduced to provide a graph-only representation:

{\small
\begin{equation}
\label{eq:fusion_fallback}
\begin{aligned}
a_i^{(m)}
&=
\frac{\exp(-u_i^{(m)})}
{\sum_{m'=1}^{M}\exp(-u_i^{(m')})+\epsilon}, 
\mathbf r_i=
\mathrm{LayerNorm}
\left(
(1-\alpha_{\mathrm{fb},i})
\sum_{m=1}^{M}a_i^{(m)}\mathbf f_i^{(m)}
+
\alpha_{\mathrm{fb},i}E_{\mathrm{struct}}(\mathbf h_{\mathrm{str},i})
\right).
\end{aligned}
\end{equation}
}

The fallback coefficient $\alpha_{\mathrm{fb},i}$ increases when the node has a high missing ratio or high average uncertainty. 
Thus, \textsc{FedMPO} can gradually shift from multimodal fusion to structure-only evidence when recovered modalities are unreliable, thereby reducing noisy semantic propagation.

\subsection{Stage 4: Task Prediction and Federated Aggregation}
\label{subsec:federated_optimization}

After obtaining the fused representation $\mathbf r_i$, each client performs a lightweight graph refinement step and applies a task-specific prediction head. 
The task loss is selected according to the downstream task, such as cross-entropy for node classification, pairwise prediction loss for link prediction, or contrastive ranking loss for modality retrieval. 
The local objective combines task learning with reconstruction, alignment, and routing regularization:
\begin{equation}
\label{eq:local_objective1}
\mathbf r'_i
=
\mathrm{LayerNorm}
\left(
\mathbf r_i+
\sigma(\mathrm{SAGEConv}(\mathbf r_i,\mathcal G))
\right)
\end{equation}
\vspace{-1em}
\begin{equation}
\label{eq:local_objective2}
\hat{\mathbf y}_i=f_{\mathrm{task}}(\mathbf r'_i), \ 
\mathcal L_k
=
\mathcal L_{\mathrm{task}}^{(k)}
+
\lambda_{\mathrm{rec}}\mathcal L_{\mathrm{rec}}^{(k)}
+
\lambda_{\mathrm{align}}\mathcal L_{\mathrm{align}}^{(k)}
+
\lambda_{\mathrm{route}}\mathcal L_{\mathrm{route}}^{(k)} .
\end{equation}
Here, $\mathcal L_{\mathrm{route}}$ contains uncertainty calibration and expert load balancing. 
This local objective ensures that the learned representation is not only task-discriminative, but also completion-aware and reliability-aware.

Finally, the server aggregates client updates. 
Standard FedAvg weights clients mainly by local data size, which can be problematic under heterogeneous modality missingness. 
A client with many nodes but unreliable recovered modalities may produce biased updates and dominate global aggregation. 
To mitigate this global aggregation bias, each client uploads lightweight reliability statistics together with model parameters, including average uncertainty $\bar u_k$, reconstruction error $\bar e_k$, and client-level missing ratio $\rho_k$. 
The server converts these statistics into reliability-aware aggregation weights:
\begin{equation}
\label{eq:relagg}
s_k
=
\exp(-\eta_u\bar u_k-\eta_e\bar e_k-\eta_\rho\rho_k),
\qquad
\omega_k
=
\frac{|\mathcal V_k|s_k}
{\sum_{j=1}^{K}|\mathcal V_j|s_j+\epsilon},
\qquad
\Theta^{t+1}
=
\sum_{k=1}^{K}\omega_k\Theta_k^{t+1}.
\end{equation}
When clients have similar modality completeness and recovery quality, this rule reduces to standard data-size-weighted FedAvg. 
Otherwise, updates from clients dominated by unreliable recovered signals are down-weighted, preventing completion errors from being repeatedly amplified across communication rounds. 
Since $s_k$ remains positive for every selected client, reliability-aware aggregation softly rescales rather than excludes client updates, preserving federated collaboration while reducing unreliable dominance.

Overall, \textsc{FedMPO} addresses topology-unaware local completion through graph-contextualized modality generation, and mitigates reliability-imbalanced global aggregation through uncertainty-guided routing and reliability-aware federated aggregation, leading to more stable optimization.

\section{Experiments}
\label{section:exps}

We conduct experiments on federated multimodal graph benchmarks under non-IID and missing-modality settings. 
Our evaluation aims to answer four questions:
\textbf{Q1} \textit{(Effectiveness)}: Does \textsc{FedMPO} outperform representative baselines across downstream tasks?
\textbf{Q2} \textit{(Interpretability)}: Do its key components contribute to robust learning?
\textbf{Q3} \textit{(Robustness)}: Is it robust to modality missingness, client heterogeneity, and hyperparameter variations?
and \textbf{Q4} \textit{(Efficiency)}: Does it remain efficient in federated training?

\textbf{Experimental setup.}
We evaluate \textsc{FedMPO} on six federated multimodal graph datasets covering node classification, link prediction, and modality retrieval. 
We report Accuracy/F1-score, AUC/AP, and Recall@5/MRR for the three tasks, respectively. 
Datasets are partitioned by Dirichlet-based non-IID splits, and both client-level and node-level modality missingness are simulated to reflect realistic cross-client modality incompleteness. 
Unless otherwise specified, the missing rate is $30\%$ and the Dirichlet concentration is $\alpha=0.5$. 
We compare with three groups of baselines: basic FGL methods, including FedAvg-Zero, FedGCN, and FedGraphSAGE; advanced FGL methods, including FedProto, FedPub, FedLAP, S2FGL, FedSPA, and FedIIH; and multimodal FL methods, including FedMVP and FedMAC. 
All methods use the same partitions, missingness protocols, and evaluation splits. 
N/A indicates that the original method is not directly applicable to the corresponding task without modifying its core learning objective; for example, prototype-based methods are not naturally defined for link prediction or retrieval.
Details are provided in Appendix~\ref{app:experiment}.

\vspace{-0.5em}
\subsection{Overall Performance}

Table~\ref{tab:main_results} summarizes the overall results. 
\textsc{FedMPO} achieves the best performance across all datasets and metrics, showing its effectiveness under modality incompleteness and federated heterogeneity. 
It consistently outperforms strong baselines on node classification, link prediction, and modality retrieval, demonstrating advantages for both graph-centric and modality-centric tasks. 
These gains verify the joint benefits of topology-aware completion, missing-aware fusion, and reliability-aware aggregation.

\begin{table*}[t]
    \centering
    \caption{Performance comparison of our proposed method against baselines across Node Classification, Link Prediction, and Modality Retrieval tasks.}
    \label{tab:main_results}
    \setlength{\tabcolsep}{1.5pt} 
    \renewcommand{\arraystretch}{1.3} 
    \resizebox{\textwidth}{!}{
    \begin{tabular}{l | c c c c | c c c c | c c c c}
        \toprule
        \multirow{3}{*}{\textbf{Methods}} & 
        \multicolumn{4}{c|}{\textbf{Node Classification}} & 
        \multicolumn{4}{c|}{\textbf{Link Prediction}} & 
        \multicolumn{4}{c}{\textbf{Modality Retrieval}} \\
        \cmidrule{2-13}
        & \multicolumn{2}{c}{Ele-fashion} & \multicolumn{2}{c|}{Grocery} 
        & \multicolumn{2}{c}{DY} & \multicolumn{2}{c|}{Bili\_Dance} 
        & \multicolumn{2}{c}{Toys} & \multicolumn{2}{c}{Flickr30k} \\
        & Acc & F1-score & Acc & F1-score 
        & AUC & AP & AUC & AP 
        & R@5 & MRR & R@5 & MRR \\
        \midrule
        
        FedAvg-Zero   & $73.92_{\pm0.41}$ & $65.73_{\pm0.46}$ & $76.21_{\pm0.51}$ & $71.18_{\pm0.57}$ & $72.31_{\pm0.62}$ & $74.18_{\pm0.53}$ & $68.41_{\pm0.61}$ & $70.18_{\pm0.64}$ & $45.18_{\pm0.79}$ & $41.12_{\pm0.76}$ & $48.31_{\pm0.68}$ & $43.48_{\pm0.81}$  \\
        
        FedGCN       & $81.96_{\pm0.35}$ & $68.44_{\pm0.31}$ & $82.53_{\pm0.39}$ & $75.39_{\pm0.46}$ & $79.37_{\pm0.49}$ & $81.21_{\pm0.41}$ & $74.77_{\pm0.49}$ & $76.11_{\pm0.47}$ & $54.41_{\pm0.64}$ & $50.87_{\pm0.61}$ & $58.18_{\pm0.56}$ & $53.72_{\pm0.64}$  \\
        FedGraphSAGE & $73.16_{\pm0.40}$ & $67.41_{\pm0.43}$ & $78.76_{\pm0.46}$ & $73.27_{\pm0.49}$ & $76.48_{\pm0.56}$ & $78.76_{\pm0.51}$ & $72.28_{\pm0.57}$ & $74.57_{\pm0.54}$ & $50.56_{\pm0.69}$ & $47.28_{\pm0.64}$ & $54.12_{\pm0.61}$ & $49.77_{\pm0.69}$  \\
        
        FedProto     & $73.61_{\pm0.38}$ & $69.38_{\pm0.39}$ & $80.12_{\pm0.41}$ & $75.61_{\pm0.46}$ & N/A & N/A & N/A & N/A & N/A & N/A & N/A & N/A   \\
        FedPub       & $\underline{84.73}_{\pm0.32}$ & $\underline{72.81}_{\pm0.54}$ & $84.57_{\pm0.36}$ & $78.36_{\pm0.43}$ & $85.38_{\pm0.41}$ & $86.76_{\pm0.39}$ & $80.86_{\pm0.44}$ & $82.37_{\pm0.41}$ & $62.46_{\pm0.54}$ & $58.07_{\pm0.49}$ & $\underline{67.38}_{\pm0.46}$ & $\underline{62.46}_{\pm0.54}$  \\
        
        FedMVP       & $77.42_{\pm0.37}$ & $70.57_{\pm0.41}$ & $82.84_{\pm0.47}$ & $75.83_{\pm0.54}$ & $83.74_{\pm0.51}$ & $84.47_{\pm0.49}$ & $78.81_{\pm0.51}$ & $80.26_{\pm0.49}$ & $58.42_{\pm0.64}$ & $54.26_{\pm0.59}$ & $62.08_{\pm0.56}$ & $57.21_{\pm0.64}$ \\
        FedMAC       & $81.28_{\pm0.41}$ & $71.14_{\pm0.39}$ & $83.47_{\pm0.44}$ & $76.48_{\pm0.49}$ & $84.48_{\pm0.46}$ & $85.58_{\pm0.41}$ & $79.37_{\pm0.47}$ & $81.12_{\pm0.44}$ & $59.78_{\pm0.61}$ & $55.42_{\pm0.54}$ & $63.22_{\pm0.51}$ & $58.37_{\pm0.59}$ \\
        FedLAP       & $77.29_{\pm0.36}$ & $70.18_{\pm0.39}$ & $\underline{85.61}_{\pm0.37}$ & $\underline{79.43}_{\pm0.41}$ & $86.17_{\pm0.41}$ & $87.73_{\pm0.34}$ & $81.76_{\pm0.41}$ & $83.48_{\pm0.39}$ & $63.46_{\pm0.49}$ & $59.18_{\pm0.47}$ & $66.76_{\pm0.44}$ & $62.08_{\pm0.49}$ \\
        S2FGL        & $74.89_{\pm0.35}$ & $69.73_{\pm0.41}$ & $84.18_{\pm0.39}$ & $77.81_{\pm0.46}$ & $85.86_{\pm0.41}$ & $87.18_{\pm0.39}$ & $81.38_{\pm0.44}$ & $83.07_{\pm0.41}$ & $\underline{64.12}_{\pm0.54}$ & $\underline{59.76}_{\pm0.51}$ & $65.77_{\pm0.47}$ & $61.18_{\pm0.54}$ \\
        FedSPA       & $75.25_{\pm0.34}$ & $70.82_{\pm0.38}$ & $85.06_{\pm0.36}$ & $78.86_{\pm0.41}$ & $\underline{87.43}_{\pm0.37}$ & $\underline{88.24}_{\pm0.31}$ & \underline{$82.26_{\pm0.39}$} & \underline{$84.18_{\pm0.37}$} & $63.76_{\pm0.47}$ & $59.48_{\pm0.44}$ & $66.47_{\pm0.41}$ & $61.76_{\pm0.47}$ \\
        FedIIH       & $75.15_{\pm0.33}$ & $70.61_{\pm0.36}$ & $84.87_{\pm0.37}$ & $78.62_{\pm0.41}$ & N/A & N/A & N/A & N/A & $63.18_{\pm0.46}$ & $58.87_{\pm0.41}$ & $66.18_{\pm0.39}$ & $61.47_{\pm0.44}$ \\
        \midrule
        
        \rowcolor[gray]{0.9} \textbf{FedMPO (Ours)}
        & $\mathbf{85.67}_{\pm0.28}$ & $\mathbf{74.82}_{\pm0.34}$ 
        & $\mathbf{86.74}_{\pm0.26}$ & $\mathbf{80.52}_{\pm0.29}$ 
        & $\mathbf{88.12}_{\pm0.24}$ & $\mathbf{89.56}_{\pm0.21}$ 
        & $\mathbf{84.18}_{\pm0.27}$ & $\mathbf{86.32}_{\pm0.26}$ 
        & $\mathbf{65.34}_{\pm0.34}$ & $\mathbf{61.18}_{\pm0.31}$ 
        & $\mathbf{68.41}_{\pm0.29}$ & $\mathbf{63.86}_{\pm0.34}$ \\
        \bottomrule
    \end{tabular}
    } 
\end{table*}

\vspace{-0.5em}
\subsection{Ablation study}
\begin{figure*}[t]
    \centering
    \includegraphics[width=\textwidth]{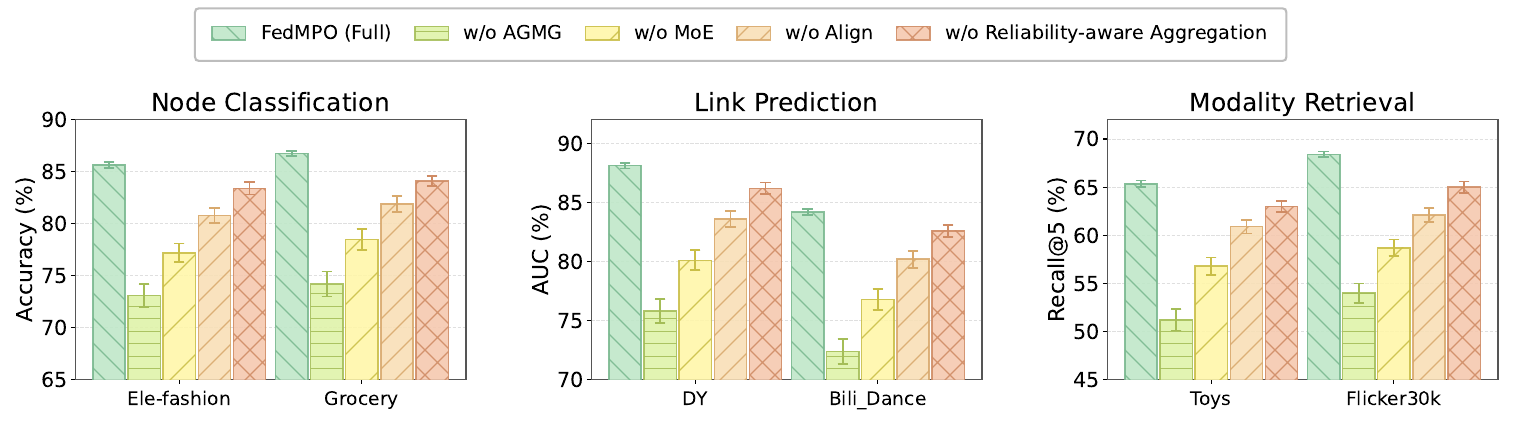}
    \caption{
    \textbf{Ablation study of \textsc{FedMPO}.}
    We compare the full model with variants removing AGMG, MoE fusion, cross-modal alignment, and reliability-aware aggregation across three downstream tasks.
    The performance drops verify the effectiveness of each component.
    }
    \label{fig:fedmpo_ablation}
\end{figure*}

We conduct module-wise ablations across node classification, link prediction, and modality retrieval. 
As shown in Fig.~\ref{fig:fedmpo_ablation}, removing any key component consistently degrades performance, confirming that \textsc{FedMPO} benefits from the cooperation of topology-aware completion, missing-aware fusion, cross-modal alignment, and reliability-aware aggregation. 
Specifically, \textit{w/o AGMG} removes the adaptive gated modality generation module, leading to the largest drop and showing that graph-structured context is crucial for missing-modality recovery. 
\textit{w/o MoE} replaces missing-aware expert fusion with simpler fusion, indicating that observed and recovered modalities should be treated according to their reliability. 
\textit{w/o Align} removes cross-modal alignment, which weakens semantic consistency among modalities. 
Finally, \textit{w/o RelAgg} replaces reliability-aware aggregation with standard FedAvg. 
Its performance drop shows that data-size-weighted aggregation may over-emphasize unreliable client updates under heterogeneous modality missingness, while \textsc{FedMPO} mitigates global aggregation bias by down-weighting clients with higher recovery uncertainty, reconstruction error, or missing ratio.

\vspace{-0.5em}
\subsection{Robustness Study}

\textbf{Missing rate sensitivity.}
Table~\ref{tab:robustness} evaluates robustness under different modality missing rates on the Ele-fashion node classification task. 
As $\eta$ increases, all methods degrade, but \textsc{FedMPO} remains more stable and exhibits a slower performance decay. 
When $\eta$ increases from $30\%$ to $70\%$, \textsc{FedMPO} drops from $85.67\%$ to $81.35\%$ in Accuracy and still outperforms the best baseline by $4.10$ percentage points. 
Thus, the performance gap increases from $0.94$ to $4.10$ points, showing that \textsc{FedMPO} is particularly advantageous under severe modality incompleteness. 
This confirms that topology-aware completion and missing-aware fusion can better preserve useful multimodal semantics when observations become sparse.

\textbf{Client heterogeneity.}
We further evaluate node classification robustness by varying the Dirichlet concentration parameter $\alpha$ while fixing the missing rate at $30\%$. 
A smaller $\alpha$ indicates stronger non-IID client partitions and more biased local data distributions. 
Under the highly heterogeneous setting $\alpha=0.1$, \textsc{FedMPO} achieves $82.45\%$ Accuracy and outperforms the best baseline by $5.65$ percentage points. 
This result indicates that reliability-aware aggregation effectively reduces the negative impact of biased and unreliable local updates, leading to more stable global optimization across heterogeneous clients.

\begin{table*}[!htbp]
    \centering
    \caption{Robustness Study on the Ele-fashion dataset. \textbf{Left:} Sensitivity to varying modality missing rates ($\eta$). \textbf{Right:} Adaptability to data heterogeneity across clients under different Dirichlet distribution parameters ($\alpha$).}
    \label{tab:robustness}
    \setlength{\tabcolsep}{2.8pt} 
    \renewcommand{\arraystretch}{1.25} 
    \resizebox{\textwidth}{!}{
    \begin{tabular}{l | c c c c c | c c c}
        \toprule
        \multirow{2}{*}{\textbf{Methods}} & 
        \multicolumn{5}{c|}{\textbf{Missing Rate Sensitivity ($\eta$) [Acc \%]}} & 
        \multicolumn{3}{c}{\textbf{Heterogeneity Analysis ($\alpha$) [Acc \%]}} \\
        \cmidrule{2-9}
        & $\eta=30\%$ & $\eta=40\%$ & $\eta=50\%$ & $\eta=60\%$ & $\eta=70\%$ 
        & $\alpha=1.0$  & $\alpha=0.5$  & $\alpha=0.1$  \\
        \midrule
        FedAvg-Zero  
        & $73.92_{\pm0.41}$ & $72.15_{\pm0.46}$ & $70.38_{\pm0.52}$ & $69.31_{\pm0.57}$ & $68.45_{\pm0.61}$ 
        & $76.80_{\pm0.38}$ & $73.92_{\pm0.41}$ & $65.12_{\pm0.68}$ \\

        FedGCN       
        & $81.96_{\pm0.35}$ & $80.24_{\pm0.41}$ & $77.85_{\pm0.49}$ & $75.96_{\pm0.54}$ & $74.15_{\pm0.58}$ 
        & $84.15_{\pm0.32}$ & $81.96_{\pm0.35}$ & $71.40_{\pm0.64}$ \\

        FedGraphSAGE 
        & $73.16_{\pm0.40}$ & $71.42_{\pm0.47}$ & $69.88_{\pm0.54}$ & $68.76_{\pm0.59}$ & $67.82_{\pm0.63}$ 
        & $75.90_{\pm0.36}$ & $73.16_{\pm0.40}$ & $64.33_{\pm0.71}$ \\

        FedProto     
        & $73.61_{\pm0.38}$ & $72.15_{\pm0.44}$ & $70.82_{\pm0.50}$ & $69.87_{\pm0.55}$ & $69.10_{\pm0.59}$ 
        & $75.40_{\pm0.35}$ & $73.61_{\pm0.38}$ & $66.85_{\pm0.66}$ \\

        FedPub       
        & $\underline{84.73}_{\pm0.32}$ & $\underline{82.55}_{\pm0.38}$ & $79.80_{\pm0.45}$ & $78.06_{\pm0.50}$ & $76.54_{\pm0.54}$ 
        & $\underline{86.10}_{\pm0.30}$ & $\underline{84.73}_{\pm0.32}$ & $75.20_{\pm0.60}$ \\

        FedMVP       
        & $77.42_{\pm0.37}$ & $75.80_{\pm0.43}$ & $73.50_{\pm0.51}$ & $69.14_{\pm0.63}$ & $65.20_{\pm0.69}$ 
        & $79.50_{\pm0.34}$ & $77.42_{\pm0.37}$ & $68.30_{\pm0.72}$ \\

        FedMAC       
        & $81.28_{\pm0.41}$ & $80.47_{\pm0.46}$ & $\underline{80.35}_{\pm0.43}$ & $\underline{78.73}_{\pm0.50}$ & $\underline{77.25}_{\pm0.55}$ 
        & $80.20_{\pm0.37}$ & $81.28_{\pm0.41}$ & $69.10_{\pm0.70}$ \\

        FedLAP       
        & $77.29_{\pm0.36}$ & $74.85_{\pm0.42}$ & $73.12_{\pm0.48}$ & $72.03_{\pm0.53}$ & $71.20_{\pm0.57}$ 
        & $78.50_{\pm0.33}$ & $77.29_{\pm0.36}$ & $\underline{76.80}_{\pm0.58}$ \\

        S2FGL        
        & $74.89_{\pm0.35}$ & $73.22_{\pm0.41}$ & $71.45_{\pm0.49}$ & $70.24_{\pm0.55}$ & $68.90_{\pm0.60}$ 
        & $76.10_{\pm0.32}$ & $74.89_{\pm0.35}$ & $65.55_{\pm0.69}$ \\

        FedSPA       
        & $75.25_{\pm0.34}$ & $73.56_{\pm0.40}$ & $71.84_{\pm0.47}$ & $70.63_{\pm0.52}$ & $69.45_{\pm0.58}$ 
        & $77.20_{\pm0.31}$ & $75.25_{\pm0.34}$ & $66.10_{\pm0.67}$ \\

        FedIIH       
        & $75.15_{\pm0.33}$ & $73.40_{\pm0.39}$ & $71.65_{\pm0.46}$ & $70.37_{\pm0.52}$ & $69.30_{\pm0.57}$ 
        & $76.90_{\pm0.30}$ & $75.15_{\pm0.33}$ & $65.80_{\pm0.66}$ \\
        \midrule

        \rowcolor[gray]{0.9} \textbf{FedMPO (Ours)} 
        & $\mathbf{85.67}_{\pm0.28}$ & $\mathbf{84.42}_{\pm0.31}$ & $\mathbf{83.15}_{\pm0.34}$ & $\mathbf{82.18}_{\pm0.36}$ & $\mathbf{81.35}_{\pm0.39}$ 
        & $\mathbf{87.20}_{\pm0.25}$ & $\mathbf{85.67}_{\pm0.28}$ & $\mathbf{82.45}_{\pm0.43}$ \\
        \bottomrule
    \end{tabular}
    }
\end{table*}

\begin{figure*}[t]
    \centering
    \begin{subfigure}[b]{0.52\linewidth}
        \centering
        \includegraphics[width=\textwidth]{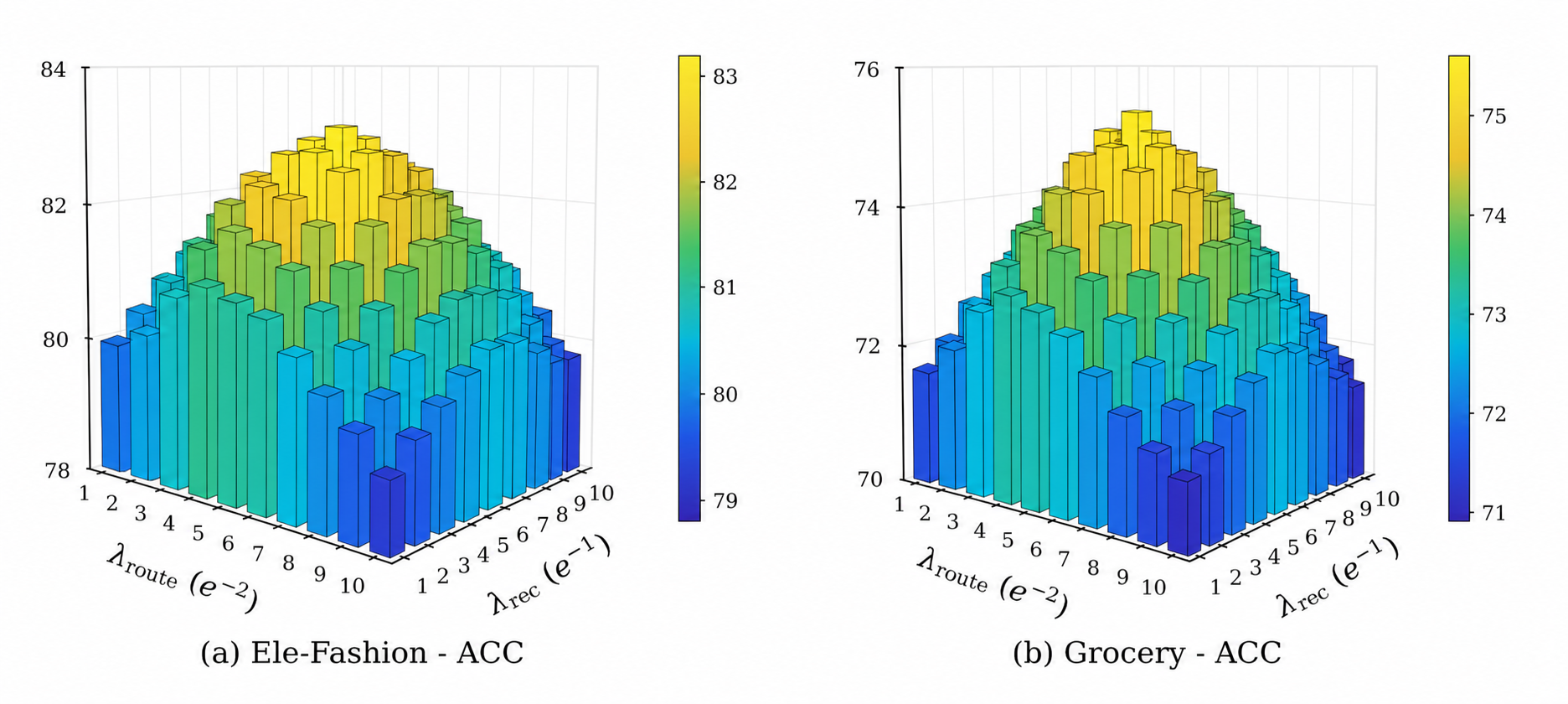}
        \caption{Sensitivity analysis of $\lambda_{\mathrm{rec}}$ and $\lambda_{\mathrm{route}}$.}
        \label{fig:hyperparameter_sensitivity_sub}
    \end{subfigure}
    \hfill
    \begin{subfigure}[b]{0.45\linewidth}
        \centering
        \includegraphics[width=\textwidth]{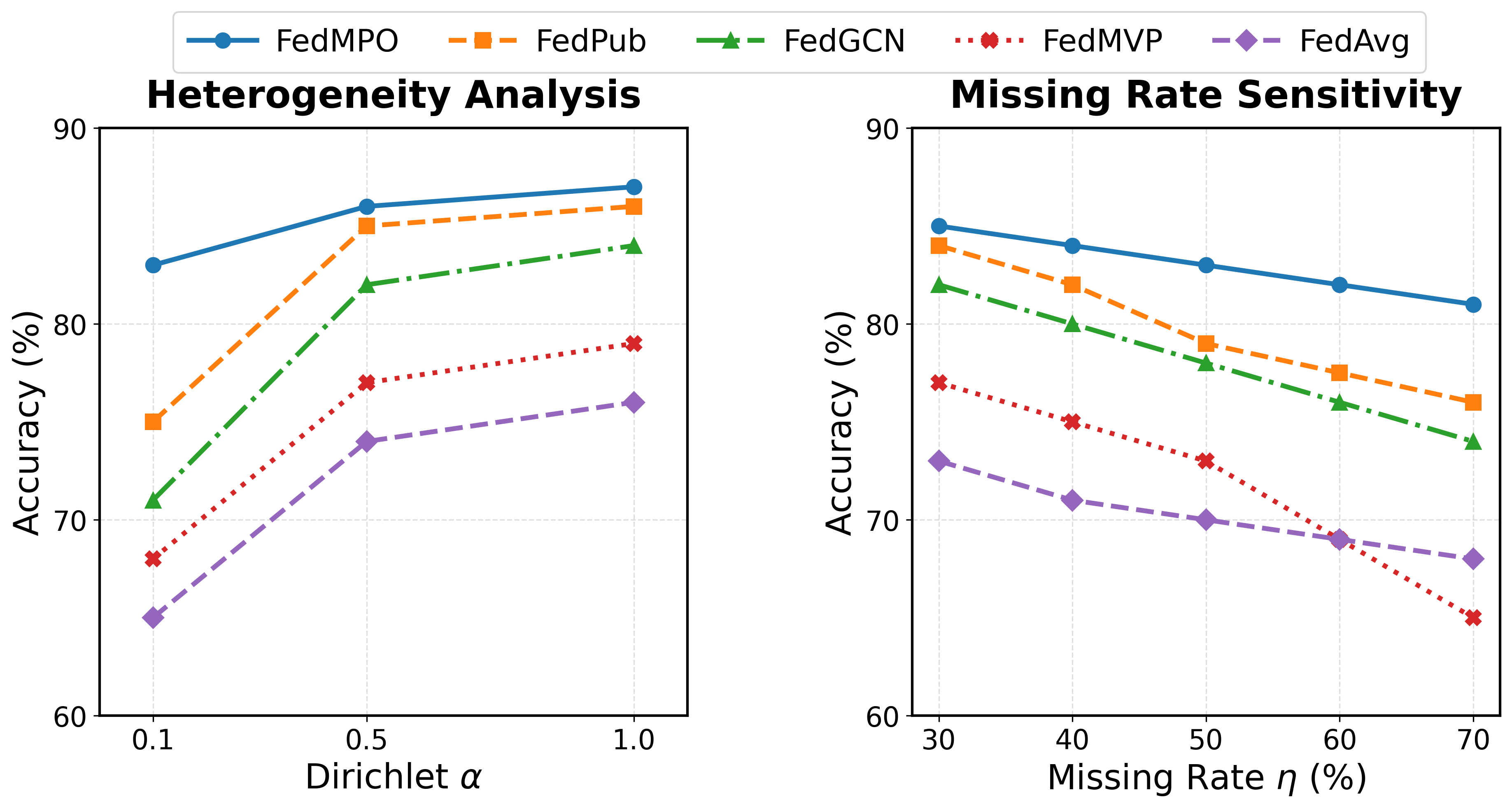}
        \caption{Robustness analysis settings.}
        \label{fig:robustness_analysis_sub}
    \end{subfigure}
    
    \caption{
    \textbf{Sensitivity and robustness analysis of \textsc{FedMPO}.}
    \textbf{(a)} Node classification ACC on Ele-Fashion and Grocery under different combinations of reconstruction coefficient $\lambda_{\mathrm{rec}}$ and routing coefficient $\lambda_{\mathrm{route}}$.
    \textbf{(b)} Robustness of \textsc{FedMPO} under different Dirichlet $\alpha$ values and modality missing rates $\eta$.
    }
    \label{fig:sensitivity_robustness}
    \vspace{-1.5em}
\end{figure*}

\textbf{Hyperparameter sensitivity.}
Fig.~\ref{fig:sensitivity_robustness}(a) reports the sensitivity of \textsc{FedMPO} to $\lambda_{\mathrm{rec}}$ and $\lambda_{\mathrm{route}}$. 
The performance remains stable across a wide range of coefficient combinations, suggesting that \textsc{FedMPO} does not rely on a narrow hyperparameter optimum. 
Fig.~\ref{fig:sensitivity_robustness}(b) further summarizes robustness trends under different missing rates and Dirichlet $\alpha$ values, confirming that \textsc{FedMPO} maintains strong performance under both modality incompleteness and client heterogeneity.

\subsection{Efficiency Study}
\textsc{FedMPO} introduces topology-aware modality generation and missing-aware expert routing, but the overhead is moderate. 
First, these modules operate on compact hidden representations rather than raw text or image inputs. 
Second, reliability-aware aggregation only requires lightweight client-level scalar statistics, including recovery uncertainty, reconstruction error, and missing ratio, whose communication cost is negligible compared with model parameters. 
Third, by filtering unreliable recovered features locally and reducing unreliable updates, \textsc{FedMPO} improves training stability under missing-modality and non-IID settings. 
Runtime and resource results are provided in Appendix~\ref{app:experiment}. 
As shown in Fig.~\ref{fig:efficiency}, \textsc{FedMPO} incurs only moderate runtime and memory overhead while remaining more efficient than several advanced FGL baselines.
Therefore, \textsc{FedMPO} achieves a favorable trade-off between accuracy, robustness, and federated training cost.

\vspace{-0.3em}
\section{Conclusion}
\label{sec:conclusion}
\vspace{-0.3em}
We presented \textsc{FedMPO}, a unified framework for missing-modality robust learning in federated multimodal graphs. 
By disentangling natural missingness from training-time masking, performing graph-enhanced cross-modal completion, and adaptively integrating observed, recovered, and structural signals through missing-aware expert fusion, \textsc{FedMPO} jointly addresses modality incompleteness, graph dependency, and cross-client heterogeneity within a single federated optimization workflow. 
Across node classification, link prediction, and modality retrieval tasks, experiments demonstrate consistent improvements over strong baselines, especially under severe missing rates and non-IID client partitions across diverse federated graph settings. 
The robustness gains further indicate that explicitly modeling recovery reliability is important when local updates are affected by heterogeneous modality availability.
These results show that reliable modality recovery, missing-aware fusion, and reliability-aware aggregation are essential for robust multimodal graph learning in federated environments. 
Limitations and future directions are discussed in Appendix~\ref{app:limitations}.

\bibliographystyle{plainnat}
\bibliography{references}

\appendix

\section{Related Works in Details}
\label{app:related}

\subsection{Centralized Multimodal Graph Learning}

Multimodal graph learning studies how to jointly model graph topology and heterogeneous node attributes, such as text and images, in order to improve structural reasoning and semantic understanding \cite{ektefaie2023multimodal,peng2024mmglsurvey}. 
Compared with unimodal graph learning, its central premise is that different modalities provide complementary semantic evidence, while graph structure offers relational context that helps propagate, refine, and align such evidence across connected nodes \cite{zhu2025mmgraph,wan2026openmag}. 
This joint modeling perspective has driven a growing line of work in recommendation, social analysis, and multimodal relational reasoning \cite{wei2019mmgcn,tao2020mgat,jia2023mhgat,he2025unigraph2}.

A first line of methods extends standard graph neural networks to multimodal node representations by designing modality-aware message passing and fusion mechanisms. 
Representative examples include MMGCN and MGAT, which perform modality-specific representation learning and then aggregate structural and cross-modal signals through graph propagation or attention-based fusion \cite{wei2019mmgcn,tao2020mgat}. 
These methods demonstrate that multimodal semantics can substantially enrich graph representations beyond structure-only or single-modality baselines. 
Subsequent work further explores more refined multimodal interaction mechanisms, heterogeneous attention, and unified multimodal embedding spaces, seeking to improve the alignment between topology and cross-modal semantics \cite{jia2023mhgat,he2025unigraph2}.

A second line of studies emphasizes the role of modality encoders and feature alignment. 
Recent benchmark efforts such as MM-GRAPH, MAGB, and OpenMAG show that multimodal graph performance is influenced not only by graph architectures, but also by how raw modalities are encoded and whether they are mapped into an aligned latent space \cite{zhu2025mmgraph,yan2024magb,wan2026openmag}. 
In particular, unified multimodal encoders such as CLIP and ImageBind often provide stronger cross-modal consistency than independent modality encoders, while graph structure can further enhance multimodal reasoning by propagating semantic information over neighborhoods \cite{radford2021clip,girdhar2023imagebind,zhu2025mmgraph,wan2026openmag}. 
These observations have made encoder choice, feature quality, and alignment quality central concerns in multimodal graph learning.

A third line of work moves beyond discriminative graph tasks and treats multimodal graphs as a more general substrate for retrieval, alignment, and generation \cite{wan2026openmag}. 
This broader view suggests that multimodal graphs are not merely graphs with extra node features, but structured systems in which topology and heterogeneous semantics mutually constrain one another. 
From this perspective, graph structure can act as a semantic scaffold, while modalities provide fine-grained content that topology alone cannot recover \cite{ektefaie2023multimodal,peng2024mmglsurvey}.

Despite this progress, existing centralized multimodal graph methods usually rely on three assumptions that do not hold in our setting. 
First, they assume that all multimodal node attributes and graph structures are available to a single training system. 
Second, they usually assume relatively complete modality observations, or at least do not explicitly model multi-granularity missingness as a first-class challenge. 
Third, even when they incorporate multimodal interaction and graph propagation effectively, they do not consider how such interactions behave under decentralized training and repeated federated aggregation. 
As a result, while centralized multimodal graph learning provides the architectural and representational foundation for our work, it does not directly address privacy-preserving distributed training with severe missing modalities.

\subsection{Federated Learning and Federated Graph Learning}

Federated learning enables multiple clients to collaboratively optimize a shared model without exposing raw local data, typically by alternating between local updates and server-side aggregation \cite{mcmahan2017fedavg}. 
This framework was originally introduced for decentralized learning on Euclidean data, and later extended to heterogeneous settings through optimization strategies such as FedProx and SCAFFOLD, which aim to reduce client drift and improve stability under non-IID data distributions \cite{li2020fedprox,karimireddy2020scaffold}. 
These methods establish the basic optimization backbone for a wide range of decentralized learning systems.

Building on this foundation, federated graph learning extends collaborative training from independent samples to graph-structured data. 
Compared with standard federated learning, graph data introduce additional challenges because local graph topology, homophily patterns, neighborhood semantics, and label distributions can vary significantly across clients. 
Recent benchmark and algorithmic studies have shown that structural heterogeneity, topology shift, and graph-specific statistical skew substantially complicate both local training and global aggregation \cite{li2025openfgl}. 
In this sense, federated graph learning is not a trivial application of standard federated optimization, but a setting in which graph structure becomes an additional source of heterogeneity.

Existing federated graph learning methods mainly focus on issues such as graph structural inconsistency, personalized aggregation, graph-specific knowledge sharing, and spectral or topology-aware regularization \cite{li2025openfgl}. 
Their central concern is how to preserve the benefits of graph representation learning under decentralized settings while mitigating the mismatch between local graph distributions. 
This literature has already demonstrated that direct adoption of standard FL algorithms can be suboptimal for graphs, and that graph-aware aggregation or personalization is often necessary.

However, most current federated graph learning methods still assume single-modality node attributes or relatively homogeneous feature spaces. 
Even when node features are high-dimensional, they are usually treated as a single vector representation rather than a set of heterogeneous modalities with different availability patterns. 
Therefore, the main challenge in standard federated graph learning is usually the mismatch of structure or labels across clients, rather than the joint interaction between graph topology and incomplete multimodal semantics.

This limitation becomes particularly important when moving to multimodal graph data. 
Once node attributes consist of multiple modalities, federated optimization must deal not only with client drift and graph heterogeneity, but also with cross-modal inconsistency, modality-dependent signal quality, and possible missingness across clients and nodes. 
Recent benchmark work such as MM-OpenFGL explicitly shows that multimodal federated graph learning constitutes a distinct problem setting rather than a straightforward extension of existing federated graph learning pipelines \cite{li2026mmopenfgl}. 
In particular, naive adaptation of standard federated learning or federated graph learning methods often fails to reconcile cross-modal semantic conflicts and cross-client structural mismatch, which motivates more targeted modeling of multimodal incompleteness and fusion reliability.

\subsection{Multimodal Federated Learning with Missing Modalities}

A related but distinct research direction is multimodal federated learning, where clients collaboratively train on multimodal data without sharing raw observations \cite{che2024fedmvp,nguyen2024fedmac,peng2024fedmm}. 
Compared with federated graph learning, the main focus here is not graph structure, but how to align, fuse, or transfer heterogeneous modalities across decentralized clients under privacy constraints. 
This line of work is especially relevant to our setting because it directly addresses incomplete modalities, modality heterogeneity, and the reliability of cross-modal information under federation.

A key theme in this literature is that multimodal heterogeneity is often more severe than the classical non-IID setting considered in standard federated learning. 
Different clients may possess different modality subsets, different encoder backbones, or different modality quality distributions. 
In such cases, the challenge is not only that data distributions differ across clients, but that the semantics represented by local observations may be structurally incomplete or asymmetric. 
This leads to a family of methods that seek to improve multimodal collaboration through prototype transfer, cross-modal aggregation, generator modules, or representation reconfiguration.

FedMVP is a representative example that leverages foundation-model-driven multimodal priors to support federated learning under incomplete modalities \cite{che2024fedmvp}. 
Its central intuition is that powerful pretrained multimodal representations can compensate for local modality absence and provide a stronger basis for cross-client collaboration. 
FedMAC, in contrast, emphasizes cross-modal aggregation and contrastive regularization for partial-modality missing scenarios, aiming to transfer information across modalities while maintaining representation consistency \cite{nguyen2024fedmac}. 
FedMM studies modality heterogeneity in federated multimodal learning from an application perspective, showing that multimodal inconsistency across clients is not merely an implementation detail but a core challenge for collaborative training \cite{peng2024fedmm}. 
FIARSE and MH-pFLID further highlight that model or modality heterogeneity requires more flexible collaboration schemes than homogeneous FL pipelines \cite{wu2024fiarse,xie2024mhpflid}.

These works provide two important insights for our problem. 
First, missing or heterogeneous modalities across clients require explicit mechanism design; they cannot be handled reliably by treating missing entries as ordinary noise or padding. 
Second, the value of recovered or transferred modality information depends on its reliability, not merely on its existence. 
This is particularly relevant for our setting, where recovered modality representations may enter graph propagation and then be repeatedly mixed through federated aggregation.

At the same time, existing multimodal federated learning methods remain insufficient for our task for two reasons. 
The first is structural: most of them are designed for non-graph multimodal data and therefore do not model neighborhood interaction, topology-conditioned completion, or graph-aware signal propagation. 
The second is reliability-related: while they address missing modalities, they typically do not study how completion errors interact with graph message passing and global aggregation over multiple communication rounds. 
For this reason, multimodal federated learning with missing modalities provides an important conceptual basis for our work, but does not fully solve missing-modality robust learning in federated multimodal graphs.

\subsection{Incomplete Multimodal Learning and Modality Completion}

Incomplete multimodal learning studies how to learn robust models when one or more modalities are absent during training or inference \cite{wu2024missingmodalitysurvey}. 
This problem has attracted increasing attention because missingness is common in practical multimodal systems, arising from sensor failure, acquisition cost, privacy restrictions, corruption, or long-term unavailability. 
Unlike standard multimodal learning, which assumes a complete set of inputs, incomplete multimodal learning must explicitly reason about how to compensate for absent information without overly trusting noisy or spurious substitutes.

A major line of methods tackles this challenge through cross-modal reconstruction or modality completion. 
The core idea is to recover a missing modality from the observed ones by exploiting cross-modal redundancy or shared latent semantics. 
These methods treat different modalities as partially predictive of each other and learn reconstruction mappings so that incomplete observations can still be projected into a semantically meaningful space \cite{wu2024missingmodalitysurvey}. 
This perspective is closely related to our generator, but in our setting the recovered representations must also be compatible with graph structure and federated optimization.

Another influential line of work focuses on shared--specific representation modeling. 
Instead of assuming that all modalities should collapse into a single common representation, these methods explicitly separate modality-shared semantics from modality-specific information \cite{wang2023sharedspecific}. 
This decomposition is useful because it acknowledges both redundancy and irreducible modality differences. 
For incomplete multimodal learning, such decomposition helps prevent trivial fusion and supports more robust transfer from observed modalities to missing ones. 
It also suggests that completion should not simply imitate a missing modality at the feature level, but should preserve the correct balance between shared semantics and modality-specific variation.

A third line of work emphasizes missing-aware fusion and reliability-aware integration. 
MMANet and related models show that the main challenge is not only to reconstruct missing modalities, but also to determine how much the model should trust recovered information relative to observed information \cite{zhao2023mmanet}. 
This perspective is particularly important in high-missing regimes, where recovered modalities may be systematically biased or uncertain. 
In such cases, blindly fusing completed features can degrade downstream performance rather than improve it.

Although incomplete multimodal learning offers important tools and insights, most of this literature remains centralized and non-graph. 
As a result, it typically ignores two mechanisms that are essential in our setting. 
The first is topology-conditioned semantics: in graphs, a node is not interpreted only through its own modalities, but also through neighborhood context. 
The second is federated error accumulation: under decentralized training, unreliable recovered information can be locally amplified through graph propagation and then globally reinforced through aggregation. 
Therefore, while incomplete multimodal learning gives a natural foundation for modality completion and missing-aware fusion, it does not by itself explain how to make such completion robust in federated multimodal graphs.

\subsection{Position of FedMPO Compared with Existing Paradigms}

FedMPO lies at the intersection of centralized multimodal graph learning, federated graph learning, multimodal federated learning, and incomplete multimodal learning, but is not reducible to any one of them. 
Compared with centralized multimodal graph learning, our setting introduces privacy-preserving distributed optimization and cross-client heterogeneity. 
Compared with federated graph learning, we explicitly study multi-granularity modality missingness rather than assuming intact or unimodal node attributes. 
Compared with multimodal federated learning under missing modalities, we further require the model to operate over graph-structured data, where node semantics are jointly shaped by multimodal observations and neighborhood context. 
Compared with incomplete multimodal learning, we must additionally consider how recovered information behaves under graph message passing and repeated federated aggregation.

This difference changes the problem formulation in an essential way. 
In centralized incomplete multimodal learning, a recovered modality mainly affects the prediction of the current sample. 
In our setting, however, a recovered modality may affect both the current node and its neighbors through graph propagation, and the resulting local parameters may further influence other clients after server aggregation. 
Therefore, the question is no longer only whether a missing modality can be reconstructed, but whether the reconstructed information is sufficiently reliable to remain useful after structural propagation and federated mixing. 
This is why completion quality, fusion reliability, and optimization stability cannot be designed independently.

FedMPO is built precisely around this observation. 
Rather than treating missing-modality recovery, graph reasoning, and federated optimization as separate components, it couples them into a single closed loop. 
Training-time mask disentanglement addresses the supervision mismatch between natural missingness and learnable reconstruction targets. 
Graph-enhanced completion injects structural context into cross-modal recovery so that missing modalities are not reconstructed in isolation. 
Missing-aware mixture-of-experts fusion explicitly models the choice among observed modalities, recovered modalities, and structural fallback, instead of assuming that all available sources should be fused indiscriminately. 
Finally, the full framework is trained under federated objectives that encourage cross-round stability rather than relying on completion alone.

In this sense, FedMPO should be viewed neither as a minor variant of incomplete multimodal learning nor as a direct adaptation of existing multimodal federated learning methods. 
Its central contribution is to elevate \emph{reliability-aware missing-modality handling under graph-structured federated optimization} into the core design principle. 
This is also the main point of departure between our work and prior paradigms: we focus not only on recovering missing modalities, but on deciding when recovered modalities should be trusted, how they should interact with graph context, and how their influence should be controlled under decentralized multimodal graph training.

\section{Methodology in Details}
\label{app:method}

This section provides additional implementation details of \textsc{FedMPO}. 
We organize the framework into four stages consistent with Sec.~3: 
feature and graph context encoding, topology-aware cross-modal generation, 
missing-aware mixture-of-experts fusion, and task prediction with reliability-aware federated aggregation.

\subsection{Stage 1: Feature and Graph Context Encoding}
\label{app:stage1}

Each client owns a private incomplete multimodal graph 
$\mathcal{G}=(\mathcal{V},\mathcal{E})$ with $N$ nodes and $M$ modalities. 
For node $v_i$ and modality $m$, the natural modality availability is denoted by 
$r_i^{(m)}\in\{0,1\}$, where $r_i^{(m)}=1$ indicates that the raw feature 
$\mathbf{x}_i^{(m)}$ is observed. 
Since naturally missing modalities have no ground-truth features, they cannot be used as direct reconstruction targets. 
Therefore, during training we only apply artificial masking to naturally observed modalities. 
Let $\bar r_i^{(m)}\in\{0,1\}$ be an artificial keep-mask sampled only when $r_i^{(m)}=1$. 
The effective visibility mask and reconstruction indicator are
\begin{equation}
\tilde r_i^{(m)} = r_i^{(m)}\bar r_i^{(m)}, 
\qquad
\Delta_i^{(m)} = r_i^{(m)}-\tilde r_i^{(m)} .
\label{eq:app_mask}
\end{equation}
Here, $\tilde r_i^{(m)}=1$ means that modality $m$ is visible to the model, while 
$\Delta_i^{(m)}=1$ means that an originally observed modality is deliberately hidden for self-supervised reconstruction. 
This design separates natural missingness from training-time masking and prevents the model from assuming that unavailable modalities are known.

For observed modalities, we map raw features into a shared latent space:
\begin{equation}
\mathbf{z}_{i,\mathrm{raw}}^{(m)} = \phi_m(\mathbf{x}_i^{(m)}),
\label{eq:app_raw}
\end{equation}
where $\phi_m(\cdot)$ is a modality-specific encoder. 
For naturally missing modalities, $\mathbf{z}_{i,\mathrm{raw}}^{(m)}$ is not computed from raw input; a learnable null token or zero vector is used only as an implementation placeholder and is excluded from reconstruction supervision.

A key difference between multimodal graphs and ordinary multimodal samples is that node semantics are shaped by both node attributes and graph neighborhoods. 
Thus, for each target modality $m$, we construct a modality-wise structural anchor from neighboring nodes whose modality-$m$ representations are visible:
\begin{equation}
\mathbf{z}_{i,\mathrm{anc}}^{(m)}
=
\frac{
\sum_{v_j\in\mathcal{N}(i)} a_{ij}\tilde r_j^{(m)}\mathbf{z}_{j,\mathrm{raw}}^{(m)}
}{
\sum_{v_j\in\mathcal{N}(i)} a_{ij}\tilde r_j^{(m)}+\epsilon
}.
\label{eq:app_anchor}
\end{equation}
Here, $a_{ij}$ is the normalized edge weight or attention coefficient, and $\epsilon$ avoids numerical instability. 
If no valid neighbor is available for modality $m$, a learnable null anchor is used. 
This anchor provides a topology-derived prior for later missing-modality generation.

The modality-wise input to the graph encoder is formed by choosing the visible raw embedding when available and the structural anchor otherwise:
\begin{equation}
\mathbf{h}_i^{(m)}
=
\mathrm{GNN}_{\theta}
\left(
\psi_m\left(
\tilde r_i^{(m)}\mathbf{z}_{i,\mathrm{raw}}^{(m)}
+
(1-\tilde r_i^{(m)})\mathbf{z}_{i,\mathrm{anc}}^{(m)}
\right),
\mathcal{G}
\right),
\label{eq:app_gnn}
\end{equation}
where $\psi_m(\cdot)$ is a lightweight modality adapter. 
For a target modality $m$, we further pool the graph-enhanced representations of other modalities into a target-exclusive context:
\begin{equation}
\mathbf{h}_{i,\setminus m}
=
\mathrm{Pool}\left(\{\mathbf{h}_i^{(m')} \mid m'\neq m\}\right).
\label{eq:app_context}
\end{equation}
This target-exclusive context prevents the generator from trivially copying the target modality during artificial masking and forces it to use complementary modalities and graph evidence.

\begin{algorithm}[htbp]
\hrule height 0.8pt
\nointerlineskip
\noindent{\fboxsep=0pt\colorbox{gray!15}{%
  \makebox[\linewidth][c]{%
    \parbox{\dimexpr\linewidth-2pt\relax}{%
      \vspace{2.5pt}%
      \footnotesize \textbf{Algorithm 1} Feature and Graph Context Encoding (Local Client)%
      \vspace{2.5pt}%
    }%
  }%
}}\par\nointerlineskip
\hrule height 0.5pt
\vspace{0.7pt}

\footnotesize
\begin{algorithmic}[1]
\REQUIRE Local graph $\mathcal{G}=(\mathcal{V},\mathcal{E})$, raw observed features $\{\mathbf{x}_i^{(m)}\}$, natural modality masks $\{r_i^{(m)}\}$.
\ENSURE Effective masks $\{\tilde r_i^{(m)}\}$, reconstruction indicators $\{\Delta_i^{(m)}\}$, raw embeddings $\{\mathbf{z}_{i,\mathrm{raw}}^{(m)}\}$, anchors $\{\mathbf{z}_{i,\mathrm{anc}}^{(m)}\}$, graph-enhanced representations $\{\mathbf{h}_i^{(m)}\}$.
\FOR{each node $v_i\in\mathcal{V}$ and modality $m$}
    \STATE // \textbf{(A) Masking and raw embedding}
    \STATE Sample artificial keep-mask $\bar r_i^{(m)}$ only if $r_i^{(m)}=1$.
    \STATE Construct $\tilde r_i^{(m)}=r_i^{(m)}\bar r_i^{(m)}$ and $\Delta_i^{(m)}=r_i^{(m)}-\tilde r_i^{(m)}$.
    \IF{$r_i^{(m)}=1$}
        \STATE Compute raw embedding $\mathbf{z}_{i,\mathrm{raw}}^{(m)}=\phi_m(\mathbf{x}_i^{(m)})$.
    \ELSE
        \STATE Use a null token or zero placeholder and exclude it from reconstruction supervision.
    \ENDIF
\ENDFOR
\FOR{each node $v_i\in\mathcal{V}$ and modality $m$}
    \STATE // \textbf{(B) Structural anchor construction}
    \STATE Compute modality-wise structural anchor:
    \STATE \hspace{\algorithmicindent} $\displaystyle
    \mathbf{z}_{i,\mathrm{anc}}^{(m)}
    =
    \frac{
    \sum_{v_j\in\mathcal{N}(i)} a_{ij}\tilde r_j^{(m)}\mathbf{z}_{j,\mathrm{raw}}^{(m)}
    }{
    \sum_{v_j\in\mathcal{N}(i)} a_{ij}\tilde r_j^{(m)}+\epsilon
    }$.
    \STATE Encode graph-enhanced modality representation:
    \STATE \hspace{\algorithmicindent} $\displaystyle
    \mathbf{h}_i^{(m)}
    =
    \mathrm{GNN}_{\theta}
    \left(
    \psi_m\left(
    \tilde r_i^{(m)}\mathbf{z}_{i,\mathrm{raw}}^{(m)}
    +
    (1-\tilde r_i^{(m)})\mathbf{z}_{i,\mathrm{anc}}^{(m)}
    \right),
    \mathcal{G}
    \right)$.
\ENDFOR
\FOR{each node $v_i\in\mathcal{V}$ and target modality $m$}
    \STATE Pool non-target modality representations to obtain $\mathbf{h}_{i,\setminus m}$.
\ENDFOR
\STATE \textbf{Return} $\{\tilde r_i^{(m)}\}$, $\{\Delta_i^{(m)}\}$, $\{\mathbf{z}_{i,\mathrm{raw}}^{(m)}\}$, $\{\mathbf{z}_{i,\mathrm{anc}}^{(m)}\}$, and $\{\mathbf{h}_i^{(m)}\}$.
\end{algorithmic}
\vspace{2pt}
\hrule height 0.5pt
\end{algorithm}

\subsection{Stage 2: Topology-aware Cross-modal Generation}
\label{app:stage2}

The second stage recovers missing modality representations. 
A naive cross-modal generator predicts a missing modality only from the remaining modalities of the same node. 
However, in multimodal graphs, a node's missing semantics may be inferred from both its own observed modalities and structurally related neighbors. 
Therefore, \textsc{FedMPO} builds a topology-aware context bank for each node and target modality.

For target modality $m$, the context bank $\mathcal{B}_i^{(m)}$ contains the current node's visible non-target modalities and neighboring visible modality representations:
\begin{equation}
\mathcal{B}_i^{(m)}
=
\left\{
\mathbf{h}_i^{(m')} \mid m'\neq m, \tilde r_i^{(m')}=1
\right\}
\cup
\left\{
\mathbf{h}_j^{(m')} \mid v_j\in\mathcal{N}(i), \tilde r_j^{(m')}=1
\right\}.
\label{eq:app_bank}
\end{equation}
During reconstruction, the target node's own modality-$m$ representation is excluded from the bank to avoid information leakage.

The query for target modality $m$ is built from the target-exclusive context, the node-level visibility pattern, and the target modality identity:
\begin{equation}
\mathbf{Q}_i^{(m)}
=
\mathbf{W}_Q
\left[
\mathbf{h}_{i,\setminus m}
\parallel
\mathbf{E}_{\mathrm{mask}}(\tilde{\mathbf{r}}_i)
\parallel
\mathbf{E}_{\mathrm{mod}}(m)
\right],
\label{eq:app_query}
\end{equation}
where $\tilde{\mathbf{r}}_i=[\tilde r_i^{(1)},\ldots,\tilde r_i^{(M)}]$, 
$\mathbf{E}_{\mathrm{mask}}(\cdot)$ embeds the missingness pattern, and 
$\mathbf{E}_{\mathrm{mod}}(m)$ is a learnable target-modality embedding. 
The cross-modal and neighborhood evidence is obtained by multi-head attention:
\begin{equation}
\mathbf{c}_i^{(m)}
=
\mathrm{MHA}\left(\mathbf{Q}_i^{(m)},\mathcal{B}_i^{(m)}\right).
\label{eq:app_attn}
\end{equation}

The final generated representation combines attention-based evidence, self context, and the structural anchor:
\begin{equation}
\hat{\mathbf{z}}_i^{(m)}
=
\gamma_t
\left(
\mathbf{g}_i^{(m)}\odot\mathbf{c}_i^{(m)}
+
(1-\mathbf{g}_i^{(m)})\odot\mathbf{W}_s\mathbf{h}_{i,\setminus m}
\right)
+
(1-\gamma_t)\mathbf{W}_a\mathbf{z}_{i,\mathrm{anc}}^{(m)} .
\label{eq:app_generation}
\end{equation}
Here, $\mathbf{g}_i^{(m)}$ is a learnable gate, and $\gamma_t$ is a round-dependent warmup factor. 
At early communication rounds, the model relies more on stable structural anchors. 
As training proceeds, it gradually shifts toward learned cross-modal generation.

The reconstruction objective is computed only on artificially masked modalities:
\begin{equation}
\mathcal{L}_{\mathrm{rec}}
=
\frac{
\sum_{i=1}^{N}\sum_{m=1}^{M}
\Delta_i^{(m)}
\left\|
\hat{\mathbf{z}}_i^{(m)}
-
\mathrm{sg}\left(\mathbf{z}_{i,\mathrm{raw}}^{(m)}\right)
\right\|_2^2
}{
\sum_{i=1}^{N}\sum_{m=1}^{M}\Delta_i^{(m)}+\epsilon
},
\label{eq:app_rec}
\end{equation}
where $\mathrm{sg}(\cdot)$ denotes the stop-gradient operation. 
Naturally missing modalities are generated for downstream learning but are not used as direct reconstruction targets.

To prevent recovered modalities from drifting away from observed modalities, we use cross-modal pairwise alignment rather than class-prototype supervision. 
Let
\begin{equation}
\tilde{\mathbf{z}}_i^{(m)}
=
\tilde r_i^{(m)}\mathbf{z}_{i,\mathrm{raw}}^{(m)}
+
(1-\tilde r_i^{(m)})\hat{\mathbf{z}}_i^{(m)} .
\label{eq:app_z_tilde}
\end{equation}
The alignment loss is
\begin{equation}
\mathcal{L}_{\mathrm{align}}
=
\frac{1}{|\Omega|}
\sum_{(i,m,m')\in\Omega}
\left[
1-
\cos
\left(
\mathbf{W}_{\mathrm{al}}\tilde{\mathbf{z}}_i^{(m)},
\mathbf{W}_{\mathrm{al}}\tilde{\mathbf{z}}_i^{(m')}
\right)
\right],
\label{eq:app_align}
\end{equation}
where $\Omega$ denotes valid modality pairs of the same node. 
This formulation does not rely on task labels and is therefore applicable to node classification, link prediction, and modality retrieval.

\begin{algorithm}[htbp]
\hrule height 0.8pt
\nointerlineskip
\noindent{\fboxsep=0pt\colorbox{gray!15}{%
  \makebox[\linewidth][c]{%
    \parbox{\dimexpr\linewidth-2pt\relax}{%
      \vspace{2.5pt}
      \footnotesize \textbf{Algorithm 2} Topology-aware Cross-modal Generation (Local Client)%
      \vspace{2.5pt}
    }%
  }%
}}\par\nointerlineskip
\hrule height 0.5pt
\vspace{0.7pt}

\footnotesize
\begin{algorithmic}[1]
\REQUIRE Graph-enhanced representations $\{\mathbf{h}_i^{(m)}\}$, raw embeddings $\{\mathbf{z}_{i,\mathrm{raw}}^{(m)}\}$, anchors $\{\mathbf{z}_{i,\mathrm{anc}}^{(m)}\}$, effective masks $\{\tilde r_i^{(m)}\}$, reconstruction indicators $\{\Delta_i^{(m)}\}$, training round $t$.
\ENSURE Generated representations $\{\hat{\mathbf{z}}_i^{(m)}\}$, reconstruction loss $\mathcal{L}_{\mathrm{rec}}$, alignment loss $\mathcal{L}_{\mathrm{align}}$.
\STATE \textbf{Hyper-parameters:} Round-dependent warmup factor $\gamma_t$.
\FOR{each node $v_i\in\mathcal{V}$ and target modality $m$}
    \STATE Build target-exclusive context $\mathbf{h}_{i,\setminus m}$ from non-target modalities.
    \STATE Build context bank $\mathcal{B}_i^{(m)}$ from visible non-target modalities and visible neighboring modality representations.
    \STATE Construct modality-specific query:
    \STATE \hspace{\algorithmicindent} $\displaystyle
    \mathbf{Q}_i^{(m)}
    =
    \mathbf{W}_Q[
    \mathbf{h}_{i,\setminus m}
    \parallel
    \mathbf{E}_{\mathrm{mask}}(\tilde{\mathbf{r}}_i)
    \parallel
    \mathbf{E}_{\mathrm{mod}}(m)
    ]$.
    \STATE Obtain attention evidence $\mathbf{c}_i^{(m)}=\mathrm{MHA}(\mathbf{Q}_i^{(m)},\mathcal{B}_i^{(m)})$.
    \STATE Generate modality representation:
    \STATE \hspace{\algorithmicindent} $\displaystyle
    \hat{\mathbf{z}}_i^{(m)}
    =
    \gamma_t
    \left(
    \mathbf{g}_i^{(m)}\odot\mathbf{c}_i^{(m)}
    +
    (1-\mathbf{g}_i^{(m)})\odot\mathbf{W}_s\mathbf{h}_{i,\setminus m}
    \right)
    +
    (1-\gamma_t)\mathbf{W}_a\mathbf{z}_{i,\mathrm{anc}}^{(m)}$.
\ENDFOR
\STATE Compute $\mathcal{L}_{\mathrm{rec}}$ only on artificially masked modalities with $\Delta_i^{(m)}=1$.
\STATE Form $\tilde{\mathbf{z}}_i^{(m)}=\tilde r_i^{(m)}\mathbf{z}_{i,\mathrm{raw}}^{(m)}+(1-\tilde r_i^{(m)})\hat{\mathbf{z}}_i^{(m)}$.
\STATE Compute $\mathcal{L}_{\mathrm{align}}$ over valid observed/recovered modality pairs $\Omega$.
\STATE \textbf{Return} $\{\hat{\mathbf{z}}_i^{(m)}\}$, $\mathcal{L}_{\mathrm{rec}}$, and $\mathcal{L}_{\mathrm{align}}$.
\end{algorithmic}
\vspace{2pt}
\hrule height 0.5pt
\end{algorithm}

\subsection{Stage 3: Missing-aware Mixture-of-Experts Fusion}
\label{app:stage3}

Although Stage~2 provides candidate representations for missing modalities, not all recovered features are equally reliable. 
Recovery is usually more reliable when the node has rich observed modalities and informative neighbors, but less reliable when both node-level observations and neighborhood evidence are sparse. 
Therefore, \textsc{FedMPO} uses uncertainty-aware expert routing to distinguish observed and recovered signals.

We first estimate modality-wise recovery uncertainty:
\begin{equation}
u_i^{(m)}
=
\begin{cases}
0, & \tilde r_i^{(m)}=1,\\
\sigma\left(
f_{\theta_u}
\left[
\hat{\mathbf{z}}_i^{(m)}
\parallel
\mathbf{h}_{i,\setminus m}
\parallel
\mathbf{z}_{i,\mathrm{anc}}^{(m)}
\right]
\right), & \tilde r_i^{(m)}=0.
\end{cases}
\label{eq:app_unc}
\end{equation}
Here, $u_i^{(m)}$ measures recovery uncertainty rather than raw-modality quality. 
Thus, visible observed modalities are assigned zero recovery uncertainty. 
The uncertainty head is calibrated on artificially masked modalities by matching uncertainty scores to normalized reconstruction errors:
\begin{equation}
\mathcal{L}_{\mathrm{unc}}
=
\frac{1}{\sum_{i,m}\Delta_i^{(m)}+\epsilon}
\sum_{i,m}
\Delta_i^{(m)}
\left(
u_i^{(m)}-\mathrm{NormErr}_i^{(m)}
\right)^2 .
\label{eq:app_unc_loss}
\end{equation}

The router considers modality visibility, recovery uncertainty, node-level missing ratio $\rho_i$, and client-level missing ratio $\rho_k$:
\begin{equation}
\left[
w_{i,\mathrm{obs}}^{(m)},
w_{i,\mathrm{rec}}^{(m)}
\right]
=
\mathrm{Softmax}
\left(
f_{\theta_r}
\left[
\tilde r_i^{(m)}
\parallel
u_i^{(m)}
\parallel
\rho_i
\parallel
\rho_k
\right]/\tau
\right).
\label{eq:app_router}
\end{equation}
The observed and recovered experts are
\begin{equation}
\mathbf{e}_{i,\mathrm{obs}}^{(m)}
=
E_{\mathrm{obs}}\left(\mathbf{z}_{i,\mathrm{raw}}^{(m)}\right),
\qquad
\mathbf{e}_{i,\mathrm{rec}}^{(m)}
=
E_{\mathrm{rec}}\left(\hat{\mathbf{z}}_i^{(m)}\right).
\label{eq:app_experts}
\end{equation}
The routed modality representation is
\begin{equation}
\mathbf{f}_i^{(m)}
=
\begin{cases}
w_{i,\mathrm{obs}}^{(m)}\mathbf{e}_{i,\mathrm{obs}}^{(m)}
+
w_{i,\mathrm{rec}}^{(m)}\mathbf{e}_{i,\mathrm{rec}}^{(m)}, 
& \tilde r_i^{(m)}=1,\\
\mathbf{e}_{i,\mathrm{rec}}^{(m)}, 
& \tilde r_i^{(m)}=0.
\end{cases}
\label{eq:app_routed}
\end{equation}

To prevent routing collapse, we use a simple load-balancing regularizer over the average routing probabilities:
\begin{equation}
\mathcal{L}_{\mathrm{bal}}
=
\sum_{q\in\{\mathrm{obs},\mathrm{rec}\}}
\left(
\bar w_q-\frac{1}{2}
\right)^2,
\qquad
\bar w_q
=
\frac{1}{NM}
\sum_{i=1}^{N}\sum_{m=1}^{M}
w_{i,q}^{(m)} .
\label{eq:app_bal}
\end{equation}
The routing loss is
\begin{equation}
\mathcal{L}_{\mathrm{route}}
=
\mathcal{L}_{\mathrm{unc}}
+
\lambda_{\mathrm{bal}}\mathcal{L}_{\mathrm{bal}} .
\label{eq:app_route_loss}
\end{equation}

After modality-level routing, modalities are fused according to their estimated reliability:
\begin{equation}
a_i^{(m)}
=
\frac{
\exp(-u_i^{(m)})
}{
\sum_{m'=1}^{M}\exp(-u_i^{(m')})+\epsilon
}.
\label{eq:app_rel_weight}
\end{equation}
For highly incomplete nodes, a structural fallback expert provides graph-only evidence. 
Let $\bar u_i=\frac{1}{M}\sum_{m=1}^{M}u_i^{(m)}$ be the average uncertainty. 
The fallback coefficient is
\begin{equation}
\alpha_{\mathrm{fb},i}
=
\sigma
\left(
f_{\mathrm{fb}}
\left[
\rho_i
\parallel
\bar u_i
\right]
\right).
\label{eq:app_fallback}
\end{equation}
The final fused node representation is
\begin{equation}
\mathbf{r}_i
=
\mathrm{LayerNorm}
\left(
(1-\alpha_{\mathrm{fb},i})
\sum_{m=1}^{M}a_i^{(m)}\mathbf{f}_i^{(m)}
+
\alpha_{\mathrm{fb},i}
E_{\mathrm{struct}}(\mathbf{h}_{\mathrm{str},i})
\right),
\label{eq:app_fusion}
\end{equation}
where $\mathbf{h}_{\mathrm{str},i}$ is a structure-only representation obtained from the local graph encoder.

\begin{algorithm}[htbp]
\hrule height 0.8pt
\nointerlineskip
\noindent{\fboxsep=0pt\colorbox{gray!15}{%
  \makebox[\linewidth][c]{%
    \parbox{\dimexpr\linewidth-2pt\relax}{%
      \vspace{2.5pt}
      \footnotesize \textbf{Algorithm 3} Missing-aware MoE Fusion (Local Client)%
      \vspace{2.5pt}
    }%
  }%
}}\par\nointerlineskip
\hrule height 0.5pt
\vspace{0.7pt}

\footnotesize
\begin{algorithmic}[1]
\REQUIRE Raw embeddings $\mathbf{z}_{i,\mathrm{raw}}^{(m)}$, generated features $\hat{\mathbf{z}}_i^{(m)}$, anchors $\mathbf{z}_{i,\mathrm{anc}}^{(m)}$, contexts $\mathbf{h}_{i,\setminus m}$, effective masks $\tilde r_i^{(m)}$, node missing ratio $\rho_i$, client missing ratio $\rho_k$.
\ENSURE Fused node representations $\{\mathbf{r}_i\}$, routing loss $\mathcal{L}_{\mathrm{route}}$, uncertainty statistics.
\FOR{each node $v_i\in\mathcal{V}$ and modality $m$}
    \STATE // \textbf{(A) Recovery uncertainty estimation}
    \IF{$\tilde r_i^{(m)}=1$}
        \STATE Assign zero recovery uncertainty $u_i^{(m)}=0$.
    \ELSE
        \STATE Estimate recovery uncertainty:
        \STATE \hspace{\algorithmicindent} $\displaystyle
        u_i^{(m)}
        =
        \sigma\left(
        f_{\theta_u}
        [
        \hat{\mathbf{z}}_i^{(m)}
        \parallel
        \mathbf{h}_{i,\setminus m}
        \parallel
        \mathbf{z}_{i,\mathrm{anc}}^{(m)}
        ]
        \right)$.
    \ENDIF
    \STATE // \textbf{(B) Expert routing}
    \STATE Compute routing weights:
    \STATE \hspace{\algorithmicindent} $\displaystyle
    [w_{i,\mathrm{obs}}^{(m)},w_{i,\mathrm{rec}}^{(m)}]
    =
    \mathrm{Softmax}
    \left(
    f_{\theta_r}
    [
    \tilde r_i^{(m)}
    \parallel
    u_i^{(m)}
    \parallel
    \rho_i
    \parallel
    \rho_k
    ]/\tau
    \right)$.
    \IF{$\tilde r_i^{(m)}=1$}
        \STATE $\mathbf{f}_i^{(m)}
        =
        w_{i,\mathrm{obs}}^{(m)}E_{\mathrm{obs}}(\mathbf{z}_{i,\mathrm{raw}}^{(m)})
        +
        w_{i,\mathrm{rec}}^{(m)}E_{\mathrm{rec}}(\hat{\mathbf{z}}_i^{(m)})$.
    \ELSE
        \STATE $\mathbf{f}_i^{(m)}=E_{\mathrm{rec}}(\hat{\mathbf{z}}_i^{(m)})$.
    \ENDIF
\ENDFOR
\FOR{each node $v_i\in\mathcal{V}$}
    \STATE // \textbf{(C) Reliability-aware fusion and structural fallback}
    \STATE Compute reliability weights $a_i^{(m)}=\frac{\exp(-u_i^{(m)})}{\sum_{m'=1}^{M}\exp(-u_i^{(m')})+\epsilon}$.
    \STATE Compute average uncertainty $\bar u_i=\frac{1}{M}\sum_{m=1}^{M}u_i^{(m)}$.
    \STATE Estimate fallback coefficient $\alpha_{\mathrm{fb},i}=\sigma(f_{\mathrm{fb}}[\rho_i\parallel\bar u_i])$.
    \STATE Fuse modality experts with structural fallback:
    \STATE \hspace{\algorithmicindent} $\displaystyle
    \mathbf{r}_i
    =
    \mathrm{LayerNorm}
    \left(
    (1-\alpha_{\mathrm{fb},i})
    \sum_{m=1}^{M}a_i^{(m)}\mathbf{f}_i^{(m)}
    +
    \alpha_{\mathrm{fb},i}E_{\mathrm{struct}}(\mathbf{h}_{\mathrm{str},i})
    \right)$.
\ENDFOR
\STATE Compute $\mathcal{L}_{\mathrm{route}}$ from uncertainty calibration and expert load balancing.
\STATE \textbf{Return} $\{\mathbf{r}_i\}$, $\mathcal{L}_{\mathrm{route}}$, and uncertainty statistics.
\end{algorithmic}
\vspace{2pt}
\hrule height 0.5pt
\end{algorithm}

\subsection{Stage 4: Task Prediction and Reliability-aware Federated Aggregation}
\label{app:stage4}

After obtaining the fused representation $\mathbf{r}_i$, each client performs a lightweight post-completion graph refinement:
\begin{equation}
\mathbf{r}'_i
=
\mathrm{LayerNorm}
\left(
\mathbf{r}_i
+
\sigma\left(\mathrm{SAGEConv}(\mathbf{R},\mathcal{G})_i\right)
\right),
\label{eq:app_refine}
\end{equation}
where $\mathbf{R}$ is the matrix of fused node representations. 
A task-specific prediction head is then applied:
\begin{equation}
\hat{\mathbf{y}}_i
=
f_{\mathrm{task}}(\mathbf{r}'_i).
\label{eq:app_pred}
\end{equation}
The task loss depends on the downstream task: cross-entropy for node classification, pairwise prediction loss for link prediction, and contrastive ranking loss for modality retrieval.

The local objective on client $k$ is
\begin{equation}
\mathcal{L}_k
=
\mathcal{L}_{\mathrm{task}}^{(k)}
+
\lambda_{\mathrm{rec}}\mathcal{L}_{\mathrm{rec}}^{(k)}
+
\lambda_{\mathrm{align}}\mathcal{L}_{\mathrm{align}}^{(k)}
+
\lambda_{\mathrm{route}}\mathcal{L}_{\mathrm{route}}^{(k)} .
\label{eq:app_local_obj}
\end{equation}
Here, $\mathcal{L}_{\mathrm{route}}^{(k)}$ contains uncertainty calibration and expert load-balancing regularization. 
This local objective jointly optimizes task prediction, self-supervised modality generation, cross-modal consistency, and missing-aware routing.

After local training, each client uploads the updated model parameters and lightweight reliability statistics to the server. 
The uploaded model parameters include the trainable parameters of the generator, fusion module, graph backbone, and task head. 
The reliability statistics include average recovery uncertainty $\bar u_k$, average reconstruction error $\bar e_k$, and client-level missing ratio $\rho_k$. 
These statistics are scalar summaries rather than raw node features, raw modality contents, recovered modality representations, or graph structures.

The server first converts the client reliability statistics into a reliability score:
\begin{equation}
s_k
=
\exp
\left(
-\eta_u\bar u_k
-\eta_e\bar e_k
-\eta_{\rho}\rho_k
\right),
\label{eq:app_score}
\end{equation}
where $\eta_u$, $\eta_e$, and $\eta_{\rho}$ are aggregation-scale coefficients controlling the effects of recovery uncertainty, reconstruction error, and client-level missingness, respectively. 
These coefficients should not be confused with the modality missing rate $\eta$ used in the robustness study.

The aggregation weight of client $k$ is then computed by combining its local graph size and reliability score:
\begin{equation}
\omega_k
=
\frac{
|\mathcal{V}_k|s_k
}{
\sum_{j\in\mathcal{S}^{t}}|\mathcal{V}_j|s_j+\epsilon
},
\label{eq:app_weight}
\end{equation}
where $\mathcal{S}^{t}$ denotes the set of selected clients at communication round $t$. 
The global model is updated as
\begin{equation}
\Theta^{t+1}
=
\sum_{k\in\mathcal{S}^{t}}
\omega_k
\Theta_k^{t+1}.
\label{eq:app_aggregation}
\end{equation}
When clients have similar modality completeness and recovery reliability, the reliability scores become similar and the rule reduces to standard data-size-weighted FedAvg. 
Otherwise, clients dominated by unreliable recovered modalities are down-weighted, preventing completion errors from being repeatedly amplified across communication rounds.

\begin{algorithm}[htbp]
\hrule height 0.8pt
\nointerlineskip
\noindent{\fboxsep=0pt\colorbox{gray!15}{%
  \makebox[\linewidth][c]{%
    \parbox{\dimexpr\linewidth-2pt\relax}{%
      \vspace{2.5pt}
      \footnotesize \textbf{Algorithm 4} Task Prediction and Reliability-aware Federated Aggregation%
      \vspace{2.5pt}
    }%
  }%
}}\par\nointerlineskip
\hrule height 0.5pt
\vspace{0.7pt}

\footnotesize
\begin{algorithmic}[1]
\REQUIRE Total communication rounds $T$, selected clients $\mathcal{S}^{t}$, initialized global model $\Theta^{0}$, reliability scales $\eta_u,\eta_e,\eta_{\rho}$.
\ENSURE Optimized global model $\Theta^{T}$.
\FOR{$t=0,1,\dots,T-1$}
    \STATE Server broadcasts global model $\Theta^{t}$ to selected clients $\mathcal{S}^{t}$.
    \FOR{each client $k\in\mathcal{S}^{t}$ in parallel}
        \STATE Initialize local model with $\Theta^{t}$.
        \STATE Execute Algorithm~1 for feature and graph context encoding.
        \STATE Execute Algorithm~2 for topology-aware cross-modal generation.
        \STATE Execute Algorithm~3 for missing-aware MoE fusion.
        \STATE // \textbf{(A) Local task prediction}
        \FOR{each node $v_i\in\mathcal{V}_k$}
            \STATE Refine fused representation:
            \STATE \hspace{\algorithmicindent} $\displaystyle
            \mathbf{r}'_i
            =
            \mathrm{LayerNorm}
            \left(
            \mathbf{r}_i
            +
            \sigma(\mathrm{SAGEConv}(\mathbf{R},\mathcal{G}_k)_i)
            \right)$.
            \STATE Predict task output $\hat{\mathbf{y}}_i=f_{\mathrm{task}}(\mathbf{r}'_i)$.
        \ENDFOR
        \STATE Compute task loss $\mathcal{L}_{\mathrm{task}}^{(k)}$ according to the downstream task.
        \STATE Compute local objective:
        \STATE \hspace{\algorithmicindent} $\displaystyle
        \mathcal{L}_k
        =
        \mathcal{L}_{\mathrm{task}}^{(k)}
        +
        \lambda_{\mathrm{rec}}\mathcal{L}_{\mathrm{rec}}^{(k)}
        +
        \lambda_{\mathrm{align}}\mathcal{L}_{\mathrm{align}}^{(k)}
        +
        \lambda_{\mathrm{route}}\mathcal{L}_{\mathrm{route}}^{(k)}$.
        \STATE Update local parameters and obtain $\Theta_k^{t+1}$.
        \STATE Compute client reliability statistics $(\bar u_k,\bar e_k,\rho_k)$.
        \STATE Upload updated parameters $\Theta_k^{t+1}$ and statistics $(\bar u_k,\bar e_k,\rho_k)$ to the server.
    \ENDFOR

    \STATE // \textbf{(B) Server-side reliability-aware aggregation}
    \FOR{each client $k\in\mathcal{S}^{t}$}
        \STATE Compute reliability score:
        \STATE \hspace{\algorithmicindent} $\displaystyle
        s_k
        =
        \exp
        \left(
        -\eta_u\bar u_k
        -\eta_e\bar e_k
        -\eta_{\rho}\rho_k
        \right)$.
        \STATE Compute aggregation weight:
        \STATE \hspace{\algorithmicindent} $\displaystyle
        \omega_k
        =
        \frac{
        |\mathcal{V}_k|s_k
        }{
        \sum_{j\in\mathcal{S}^{t}}|\mathcal{V}_j|s_j+\epsilon
        }$.
    \ENDFOR
    \STATE Update global model:
    \STATE \hspace{\algorithmicindent} $\displaystyle
    \Theta^{t+1}
    =
    \sum_{k\in\mathcal{S}^{t}}
    \omega_k\Theta_k^{t+1}$.
\ENDFOR
\STATE \textbf{Return} $\Theta^{T}$.
\end{algorithmic}
\vspace{2pt}
\hrule height 0.5pt
\end{algorithm}

\section{Experiment in Details}
\label{app:experiment}

\subsection{Computation Resource}

All experiments were conducted on distributed container nodes of the AutoDL cloud computing platform. 
Table~\ref{tab:computing} summarizes the machine configuration and software stack used in our experiments. 
The environment provides sufficient GPU memory for multimodal feature processing and federated graph training.

\begin{table}[h]
\caption{Compute environment used for all experiments.}
\label{tab:computing}
\centering
\begin{tabular}{ll}
\toprule
\rowcolor[gray]{0.9} \textbf{Component} & \textbf{Configuration} \\
\midrule
Server & AutoDL Cloud Computing Platform \\
CPU & $2 \times$ Intel(R) Xeon(R) Platinum 8470Q @ 3.80GHz (104 cores, 208 threads total) \\
Memory & 1.0 TiB RAM \\
GPU & $2 \times$ NVIDIA RTX PRO 6000 Blackwell (96 GiB each); Driver: 590.44.01 \\
OS / Kernel & Ubuntu 22.04.5 LTS; Linux 5.15.0-78-generic \\
CUDA & Runtime 12.8; Driver Version: 590.44 \\
Python & Python 3.12.3 \\
DL stack & PyTorch 2.8.0, PyG 2.7.0, scikit-learn 1.6.0 \\
\bottomrule
\end{tabular}
\end{table}

\subsection{Datasets}

We evaluate \textsc{FedMPO} on six multimodal graph datasets covering three downstream tasks: node classification, link prediction, and modality retrieval. 
For all datasets, image or video modalities are encoded into 512-dimensional features, while textual modalities are encoded into 768-dimensional features. 
Dataset statistics are summarized in Table~\ref{tab:datasets}.

\textbf{Ele-fashion} is a heterogeneous product graph constructed by merging Amazon Electronics and Fashion categories. 
Nodes represent products, and edges are derived from cross-category co-purchasing relations. 
Textual features are extracted from product titles, descriptions, and specifications, while visual features are extracted from product images. 
This dataset is used for node classification, where labels correspond to high-level product categories.

\textbf{Grocery} is sourced from Amazon Grocery and Gourmet Food. 
Nodes represent food or household products, and edges are built from co-purchasing relations. 
Textual attributes are extracted from product titles and descriptions, while visual attributes are extracted from packaging images. 
This dataset is used for node classification with fine-grained product categories.

\textbf{DY} is sourced from a short-video platform. 
Nodes represent short videos, and edges are constructed from co-interaction relations, such as videos liked or watched by similar users. 
Textual features are extracted from captions and hashtags, while visual features are obtained from sampled video frames. 
This dataset is used for link prediction.

\textbf{Bili\_Dance} is collected from the dance category of Bilibili. 
Nodes represent dance videos, and edges encode co-viewing or sequential watching relations among users. 
Textual attributes are derived from video descriptions and tags, while visual features are extracted from keyframes. 
This dataset is used for link prediction, evaluating whether the learned representations can recover potential structural relations under missing modalities.

\textbf{Toys} originates from Amazon Toys and Games. 
Nodes represent toy products, and edges are derived from co-purchasing relations. 
Textual features are extracted from product descriptions and age recommendations, while visual features are extracted from product images. 
This dataset is used for modality retrieval.

\textbf{Flickr30k} is an image-text dataset for multimodal alignment. 
We construct a multimodal graph where nodes correspond to images or textual descriptions, and edges reflect semantic image-text associations. 
This dataset is used for modality retrieval, evaluating cross-modal matching between visual and textual representations.

\begin{table}[h]
\caption{Statistics of the experimental datasets.}
\label{tab:datasets}
\centering
\begin{tabular}{l|c|c|c|c}
\hline
\rowcolor[gray]{0.9} \textbf{Dataset} & \textbf{Modalities} & \textbf{Feature Dimensions} & \textbf{Samples / Nodes} & \textbf{Task} \\
\hline
Ele-fashion & Image + Text & 512 + 768 & $\sim$10,500 & Node Classification \\
Grocery & Image + Text & 512 + 768 & $\sim$8,200 & Node Classification \\
DY & Video + Text & 512 + 768 & $\sim$12,000 & Link Prediction \\
Bili\_Dance & Video + Text & 512 + 768 & $\sim$9,800 & Link Prediction \\
Toys & Image + Text & 512 + 768 & $\sim$5,400 & Modality Retrieval \\
Flickr30k & Image + Text & 512 + 768 & 31,783 & Modality Retrieval \\
\hline
\end{tabular}
\end{table}

\subsection{Baseline Models}

We compare \textsc{FedMPO} with three groups of baselines: basic federated graph learning methods, advanced federated graph learning methods, and multimodal federated learning methods. 
For fair comparison, all baselines use the same multimodal features, client partitions, missing-modality protocols, and evaluation splits.

\paragraph{Basic federated graph learning baselines.}

\textbf{FedAvg-Zero} adapts standard FedAvg to incomplete multimodal graphs by zero-padding missing modality features before local training. 
This baseline evaluates whether simple missing-feature filling is sufficient under federated multimodal graph settings.

\textbf{FedGCN} extends graph convolutional networks to federated training. 
It models local graph topology but does not explicitly recover or distinguish missing multimodal features.

\textbf{FedGraphSAGE} applies the inductive GraphSAGE architecture in a federated setting. 
Although scalable for graph representation learning, it assumes directly usable node features and is vulnerable when modality observations are incomplete.

\paragraph{Advanced federated graph learning baselines.}

\textbf{FedProto} mitigates statistical heterogeneity by sharing class-level prototypes rather than only model parameters. 
However, under missing modalities, locally estimated prototypes can become biased by incomplete feature observations.

\textbf{FedPub} performs personalized subgraph federated learning by estimating client similarities and adaptively aggregating model updates. 
It handles subgraph heterogeneity but does not explicitly model modality missingness or completion reliability.

\textbf{FedLAP} uses Laplacian-based graph regularization to improve federated graph learning under structural heterogeneity. 
It focuses on graph structure but lacks cross-modal recovery mechanisms.

\textbf{S2FGL} combines spatial and spectral information to reduce client drift in federated graph learning. 
It is designed for graph distribution shifts but does not explicitly address incomplete multimodal attributes.

\textbf{FedSPA} addresses homophily heterogeneity in federated graph learning through structure-aware propagation and aggregation. 
Its design mainly targets topology heterogeneity rather than missing multimodal features.

\textbf{FedIIH} models inter-client and intra-client heterogeneity in federated subgraphs. 
Although effective for heterogeneous graph distributions, it does not explicitly recover missing modalities or estimate completion reliability.

\paragraph{Multimodal federated learning baselines.}

\textbf{FedMVP} leverages multimodal priors for federated learning with incomplete modalities. 
It is effective for non-graph multimodal data, but it does not explicitly use graph neighborhoods for topology-aware modality recovery.

\textbf{FedMAC} addresses partial-modality missingness with cross-modal aggregation and contrastive regularization. 
However, it treats samples independently and does not model graph-structured dependencies during completion.

\subsection{Downstream Tasks}

We evaluate \textsc{FedMPO} on graph-centric and modality-centric downstream tasks. 
Graph-centric tasks include node classification and link prediction, while the modality-centric task is modality retrieval. 
These tasks jointly evaluate whether \textsc{FedMPO} can learn robust representations under incomplete modalities, graph heterogeneity, and federated non-IID partitions.

\paragraph{Node Classification.}
Node classification evaluates whether the learned node representations are discriminative under incomplete multimodal inputs. 
Given node embeddings produced by the local graph model, a task-specific prediction head outputs class probabilities. 
We evaluate node classification on Ele-fashion and Grocery using Accuracy and F1-score.

\paragraph{Link Prediction.}
Link prediction evaluates whether the model can infer missing or potential edges from multimodal graph representations. 
For each candidate node pair, the model computes a similarity or scoring function based on the learned embeddings. 
Positive edges are sampled from observed graph links, while negative edges are generated by negative sampling. 
We evaluate link prediction on DY and Bili\_Dance using AUC and AP.

\paragraph{Modality Retrieval.}
Modality retrieval evaluates cross-modal alignment under incomplete modalities. 
Given a query from one modality, the model ranks candidate instances from another modality in the shared representation space. 
This task directly reflects whether recovered and observed modalities are aligned in a retrieval setting. 
We evaluate modality retrieval on Toys and Flickr30k using Recall@K and MRR.

\subsection{Evaluation Metrics}

We use standard evaluation metrics for the three downstream tasks.

\textbf{Accuracy (Acc).}
For node classification, Accuracy measures the proportion of correctly classified nodes:
\begin{equation}
\mathrm{Acc}
=
\frac{1}{N}
\sum_{i=1}^{N}
\mathbb{I}(\hat{y}_i=y_i),
\end{equation}
where $N$ is the number of evaluated nodes, $y_i$ is the ground-truth label, and $\hat{y}_i$ is the predicted label.

\textbf{F1-score.}
F1-score is the harmonic mean of Precision and Recall:
\begin{equation}
\mathrm{F1}
=
2\cdot
\frac{\mathrm{Precision}\cdot \mathrm{Recall}}
{\mathrm{Precision}+\mathrm{Recall}}.
\end{equation}
It complements Accuracy when class distributions are imbalanced.

\textbf{Area Under the ROC Curve (AUC).}
For link prediction, AUC evaluates whether positive edges receive higher scores than negative edges:
\begin{equation}
\mathrm{AUC}
=
\frac{
\sum_{i\in\mathcal{P}}\mathrm{rank}_i
-
\frac{|\mathcal{P}|(|\mathcal{P}|+1)}{2}
}{
|\mathcal{P}|\times|\mathcal{N}|
},
\end{equation}
where $\mathcal{P}$ and $\mathcal{N}$ denote positive and negative edge sets.

\textbf{Average Precision (AP).}
AP summarizes the precision-recall curve:
\begin{equation}
\mathrm{AP}
=
\sum_n
(R_n-R_{n-1})P_n,
\end{equation}
where $P_n$ and $R_n$ denote the precision and recall at the $n$-th threshold.

\textbf{Recall@K (R@K).}
For modality retrieval, Recall@K measures whether the correct target appears in the top-$K$ retrieved candidates:
\begin{equation}
\mathrm{R@K}
=
\frac{1}{|Q|}
\sum_{i=1}^{|Q|}
\mathbb{I}(\mathrm{rank}_i\leq K),
\end{equation}
where $Q$ is the query set.

\textbf{Mean Reciprocal Rank (MRR).}
MRR evaluates the rank position of the first correct retrieved result:
\begin{equation}
\mathrm{MRR}
=
\frac{1}{|Q|}
\sum_{i=1}^{|Q|}
\frac{1}{\mathrm{rank}_i}.
\end{equation}

\subsection{Further Experimental Analysis and Discussions}

\subsubsection{\normalfont\textsc{Additional Ablation Analysis}}

\begin{figure}[t]
    \centering
    \includegraphics[width=\linewidth]{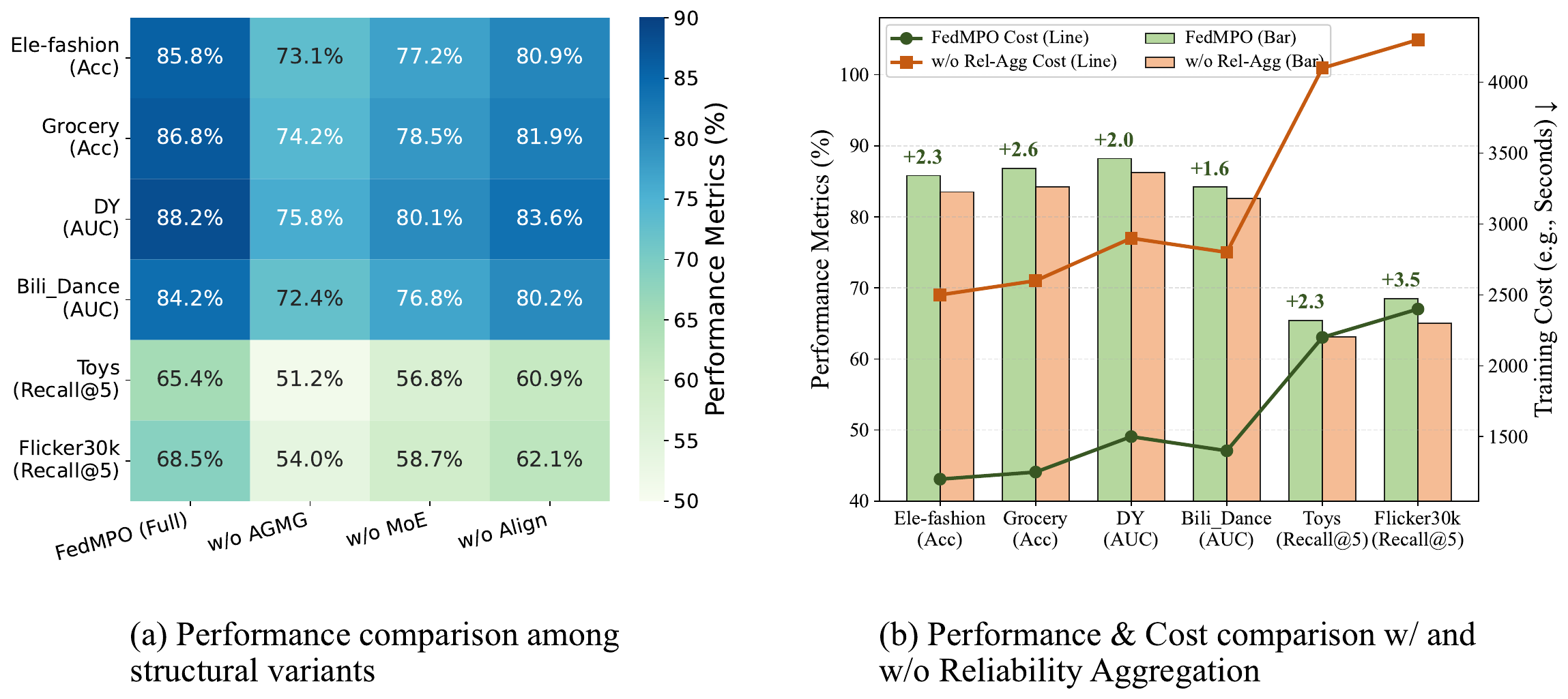}
    \caption{
    \textbf{Ablation study of \textsc{FedMPO}. } 
    (a) Consistent performance drops in variants (w/o AGMG, MoE, Align) prove the necessity of each structural design. (b) Comparisons w/ and w/o RelAgg confirm that reliability-aware aggregation is critical for stabilizing performance and reducing training costs.
    }
    \label{fig:fedmpo_ablation_appendix}
\end{figure}

We provide a more detailed ablation analysis in Fig.~\ref{fig:fedmpo_ablation_appendix}. 
The left panel compares the full \textsc{FedMPO} model with three structural variants: w/o AGMG, w/o MoE, and w/o Align. 
Specifically, w/o AGMG removes the topology-aware cross-modal generation module, w/o MoE replaces missing-aware expert routing with simple fusion, and w/o Align removes the cross-modal alignment objective. 
Across node classification, link prediction, and modality retrieval datasets, all ablated variants lead to consistent performance degradation, verifying that each component contributes to robust federated multimodal graph learning. 
Among them, removing AGMG causes the largest drop in most cases, indicating that topology-aware modality completion is essential for recovering reliable missing representations. 
The degradation of w/o MoE further shows that simply fusing observed and recovered modalities is insufficient, since recovered features may contain task-irrelevant or noisy semantics under severe modality missingness. 
The performance drop of w/o Align also confirms that cross-modal consistency regularization helps reduce semantic mismatch between observed and generated modality representations.

The right panel further compares \textsc{FedMPO} with w/o RelAgg, where reliability-aware aggregation is replaced by standard data-size-weighted FedAvg. 
\textsc{FedMPO} consistently achieves better performance across all six datasets, with gains ranging from 1.6 to 3.5 percentage points. 
This demonstrates that aggregation based only on local data size can over-emphasize unreliable client updates when clients have heterogeneous modality availability and recovery quality. 
By incorporating client-level uncertainty, reconstruction error, and missing ratio into aggregation weights, reliability-aware aggregation improves the robustness of the global model. 
Meanwhile, the cost comparison suggests that \textsc{FedMPO} maintains a favorable performance--cost trade-off, since more reliable aggregation can stabilize federated optimization without introducing heavy communication overhead.

\subsubsection{\normalfont\textsc{MoE Router Diagnostics and Expert Behavior}}

To analyze the behavior of the missing-aware MoE module, we track routing probabilities, recovery uncertainty, expert output norms, and the similarity between recovered representations and structure-only fallback representations. 
At early communication rounds, recovered features are less reliable, and the router assigns relatively higher weights to the structural fallback. 
As the topology-aware generator becomes better trained, the router gradually increases the contribution of recovered and observed modality experts. 
This trend suggests that the router learns to adaptively adjust the reliance on observed, recovered, and structural signals according to recovery reliability.

\subsubsection{\normalfont\textsc{Extended Robustness to Varying Missing Rates}}

Beyond the main robustness study on node classification, we further evaluate \textsc{FedMPO} under missing rates ranging from 30\% to 70\% on link prediction and modality retrieval tasks. 
As missingness increases, all methods degrade, but \textsc{FedMPO} shows a slower performance drop. 
This is because AGMG exploits neighborhood context to provide topology-aware recovery, while the missing-aware router reduces the influence of uncertain recovered modalities. 
These results support the robustness of \textsc{FedMPO} under severe modality incompleteness.

\subsubsection{\normalfont\textsc{Round-wise Convergence}}

We also compare round-wise convergence across methods. 
\textsc{FedMPO} reaches a stable performance region in fewer communication rounds than baselines that either ignore missing modalities or rely on standard aggregation. 
The improvement mainly comes from two factors: unreliable recovered signals are filtered during local routing, and unreliable client updates are down-weighted during server aggregation. 
Therefore, \textsc{FedMPO} improves not only final performance but also optimization stability under missing-modality and non-IID settings.

\subsubsection{\normalfont\textsc{Runtime and Resource Usage Study}}

We report system-level efficiency in Fig.~\ref{fig:efficiency} to evaluate the additional cost introduced by topology-aware generation and missing-aware routing. 
Compared with FedAvg-Zero, \textsc{FedMPO} introduces computational and memory overhead (as shown in Fig. \ref{fig:runtime} and \ref{fig:memory}) due to AGMG, uncertainty estimation, and MoE routing. 
However, these modules operate on compact hidden representations rather than raw image or text inputs. 
Moreover, reliability-aware aggregation only requires lightweight scalar statistics, including average recovery uncertainty, reconstruction error, and missing ratio. 
As a result, the additional communication cost is small compared with transmitting model parameters. 
Overall, as demonstrated by its high throughput relative to other baselines,\textsc{FedMPO} provides a favorable trade-off between performance improvement and computational overhead.

\definecolor{c1D}{HTML}{3D7CB4} \definecolor{c1M}{HTML}{8EBAE5} \definecolor{c1L}{HTML}{D1E5F0} 
\definecolor{c2D}{HTML}{64A071} \definecolor{c2L}{HTML}{A6D0AA} 
\definecolor{c3D}{HTML}{C54576} \definecolor{c3L}{HTML}{DFA6C1} 
\definecolor{c4D}{HTML}{9078B4} \definecolor{c4L}{HTML}{CFC1DF} 
\definecolor{c5D}{HTML}{D15C53} \definecolor{c5L}{HTML}{F6A78D} 
\definecolor{oursD}{HTML}{8562A2}  
\definecolor{oursM1}{HTML}{9F82B5} 
\definecolor{oursM2}{HTML}{B9A2C8} 
\definecolor{oursL}{HTML}{E2D9EB}  

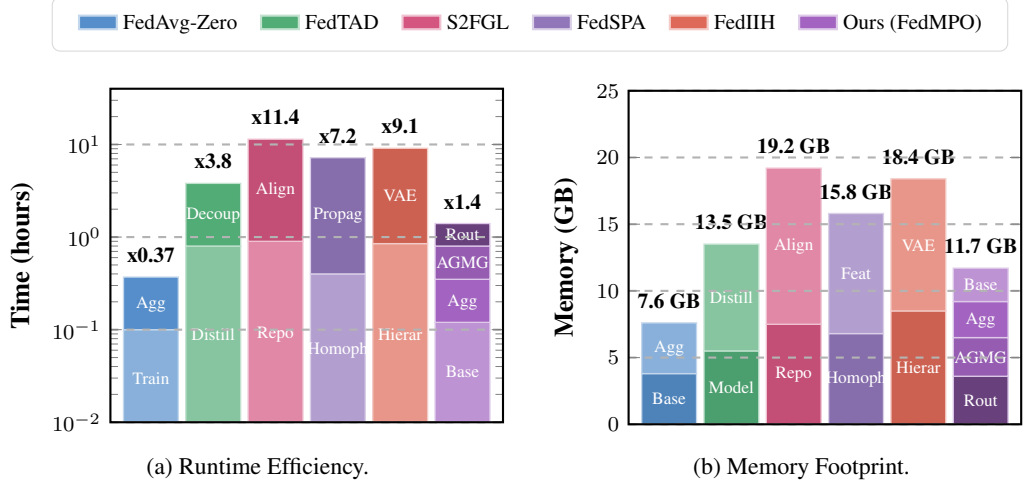
\begin{figure*}[htbp]
    \definecolor{c1}{RGB}{85, 142, 202}  
    \definecolor{c2}{RGB}{82, 172, 118}  
    \definecolor{c3}{RGB}{212, 85, 128}  
    \definecolor{c4}{RGB}{146, 118, 186} 
    \definecolor{c5}{RGB}{222, 105, 88}  
    \definecolor{c6}{RGB}{162, 100, 192} 
    
    \colorlet{c1D}{c1!92!black} \colorlet{c1M}{c1} \colorlet{c1L}{c1!70!white} \colorlet{c1O}{c1!35!white}
    \colorlet{c2D}{c2!92!black} \colorlet{c2M}{c2} \colorlet{c2L}{c2!70!white} \colorlet{c2O}{c2!35!white}
    \colorlet{c3D}{c3!92!black} \colorlet{c3M}{c3} \colorlet{c3L}{c3!70!white} \colorlet{c3O}{c3!35!white}
    \colorlet{c4D}{c4!92!black} \colorlet{c4M}{c4} \colorlet{c4L}{c4!70!white} \colorlet{c4O}{c4!35!white}
    \colorlet{c5D}{c5!92!black} \colorlet{c5M}{c5} \colorlet{c5L}{c5!70!white} \colorlet{c5O}{c5!35!white}
    \colorlet{c6D}{c6!92!black} \colorlet{c6M}{c6} \colorlet{c6L}{c6!70!white} \colorlet{c6O}{c6!35!white}

    \tikzstyle{bt}=[text=white, font=\fontsize{6.5pt}{7.5pt}\selectfont\rmfamily, align=center]

    \begin{center}
        \begin{tikzpicture}
            \setlength{\tabcolsep}{1.5mm}
            \node[draw=black!15, fill=white, rounded corners=3pt, inner sep=1.2ex] at (0,0) {
                \footnotesize\rmfamily 
                \begin{tabular}{cccccc}
                    \tikz[baseline=0cm]{\filldraw[fill=c1M, draw=c1O, line width=0.8pt, sharp corners] (0,0) rectangle (0.5cm,0.18cm);} FedAvg-Zero &
                    \tikz[baseline=0cm]{\filldraw[fill=c2M, draw=c2O, line width=0.8pt, sharp corners] (0,0) rectangle (0.5cm,0.18cm);} FedTAD &
                    \tikz[baseline=0cm]{\filldraw[fill=c3M, draw=c3O, line width=0.8pt, sharp corners] (0,0) rectangle (0.5cm,0.18cm);} S2FGL &
                    \tikz[baseline=0cm]{\filldraw[fill=c4M, draw=c4O, line width=0.8pt, sharp corners] (0,0) rectangle (0.5cm,0.18cm);} FedSPA &
                    \tikz[baseline=0cm]{\filldraw[fill=c5M, draw=c5O, line width=0.8pt, sharp corners] (0,0) rectangle (0.5cm,0.18cm);} FedIIH &
                    \tikz[baseline=0cm]{\filldraw[fill=c6M, draw=c6O, line width=0.8pt, sharp corners] (0,0) rectangle (0.5cm,0.18cm);} Ours (FedMPO)
                \end{tabular}
            };
        \end{tikzpicture}
    \end{center}
    
    \vspace{0.1cm}

    \noindent 
    \begin{subfigure}{0.485\textwidth} 
        \begin{tikzpicture}
            \begin{axis}[
                width=\linewidth, 
                height=6.0cm, 
                ymode=log, ymin=0.01, ymax=40, xmin=0.35, xmax=6.65, 
                ylabel={Time (hours)}, 
                label style={font=\normalsize\rmfamily\bfseries}, 
                tick label style={font=\footnotesize\rmfamily}, 
                xtick=\empty, 
                ymajorgrids=true, 
                grid style={dashed, gray!60, line width=0.8pt}, 
                axis on top=true, 
                axis line style={line width=0.8pt},
            ]
            
            \filldraw[fill=c1L, draw=c1O, line width=0.5pt] (axis cs:0.56,0.01) rectangle (axis cs:1.44,0.1); \node[bt] at (axis cs:1,0.03) {Train};
            \filldraw[fill=c1M, draw=c1O, line width=0.5pt] (axis cs:0.56,0.1) rectangle (axis cs:1.44,0.37); \node[bt] at (axis cs:1,0.19) {Agg};
            \node[font=\footnotesize\rmfamily\bfseries, color=black, anchor=south, yshift=2pt] at (axis cs:1,0.37) {x0.37};

            \filldraw[fill=c2L, draw=c2O, line width=0.5pt] (axis cs:1.56,0.01) rectangle (axis cs:2.44,0.8); \node[bt] at (axis cs:2,0.09) {Distill};
            \filldraw[fill=c2M, draw=c2O, line width=0.5pt] (axis cs:1.56,0.8) rectangle (axis cs:2.44,3.8);  \node[bt] at (axis cs:2,1.74) {Decoup};
            \node[font=\footnotesize\rmfamily\bfseries, color=black, anchor=south, yshift=2pt] at (axis cs:2,3.8) {x3.8};

            \filldraw[fill=c3L, draw=c3O, line width=0.5pt] (axis cs:2.56,0.01) rectangle (axis cs:3.44,0.9); \node[bt] at (axis cs:3,0.09) {Repo};
            \filldraw[fill=c3D, draw=c3O, line width=0.5pt] (axis cs:2.56,0.9) rectangle (axis cs:3.44,11.4); \node[bt] at (axis cs:3,3.2) {Align};
            \node[font=\footnotesize\rmfamily\bfseries, color=black, anchor=south, yshift=2pt] at (axis cs:3,11.4) {x11.4};

            \filldraw[fill=c4L, draw=c4O, line width=0.5pt] (axis cs:3.56,0.01) rectangle (axis cs:4.44,0.4); \node[bt] at (axis cs:4,0.06) {Homoph};
            \filldraw[fill=c4D, draw=c4O, line width=0.5pt] (axis cs:3.56,0.4) rectangle (axis cs:4.44,7.2);  \node[bt] at (axis cs:4,1.7) {Propag};
            \node[font=\footnotesize\rmfamily\bfseries, color=black, anchor=south, yshift=2pt] at (axis cs:4,7.2) {x7.2};

            \filldraw[fill=c5L, draw=c5O, line width=0.5pt] (axis cs:4.56,0.01) rectangle (axis cs:5.44,0.85); \node[bt] at (axis cs:5,0.09) {Hierar};
            \filldraw[fill=c5D, draw=c5O, line width=0.5pt] (axis cs:4.56,0.85) rectangle (axis cs:5.44,9.1);  \node[bt] at (axis cs:5,2.8) {VAE};
            \node[font=\footnotesize\rmfamily\bfseries, color=black, anchor=south, yshift=2pt] at (axis cs:5,9.1) {x9.1};

            \filldraw[fill=c6L, draw=c6O, line width=0.5pt] (axis cs:5.56,0.01) rectangle (axis cs:6.44,0.12); \node[bt] at (axis cs:6,0.035) {Base};
            \filldraw[fill=c6M, draw=c6O, line width=0.5pt] (axis cs:5.56,0.12) rectangle (axis cs:6.44,0.35); \node[bt] at (axis cs:6,0.2) {Agg};
            \filldraw[fill=c6D, draw=c6O, line width=0.5pt] (axis cs:5.56,0.35) rectangle (axis cs:6.44,0.8);  \node[bt] at (axis cs:6,0.53) {AGMG};
            \filldraw[fill=c6D!70!black, draw=c6O, line width=0.5pt] (axis cs:5.56,0.8) rectangle (axis cs:6.44,1.4); \node[bt] at (axis cs:6,1.06) {Rout};
            \node[font=\footnotesize\rmfamily\bfseries, color=black, anchor=south, yshift=2pt] at (axis cs:6,1.4) {x1.4};

            \end{axis}
        \end{tikzpicture}
        \caption{Runtime Efficiency.}
        \label{fig:runtime} %
    \end{subfigure}%
    \hfill%
    \begin{subfigure}{0.485\textwidth}
        \begin{tikzpicture}
            \begin{axis}[
                width=\linewidth, 
                height=6.0cm,
                ymin=0, ymax=25, xmin=0.35, xmax=6.65,
                ytick={0,5,10,15,20,25}, 
                ylabel={Memory (GB)}, 
                label style={font=\normalsize\rmfamily\bfseries}, 
                tick label style={font=\footnotesize\rmfamily}, 
                xtick=\empty, 
                ymajorgrids=true, 
                grid style={dashed, gray!60, line width=0.8pt}, 
                axis on top=true, 
                axis line style={line width=0.8pt},
            ]

            \filldraw[fill=c1D, draw=c1O, line width=0.5pt] (axis cs:0.56,0) rectangle (axis cs:1.44,3.8);   \node[bt] at (axis cs:1,1.9) {Base};
            \filldraw[fill=c1L, draw=c1O, line width=0.5pt] (axis cs:0.56,3.8) rectangle (axis cs:1.44,7.6); \node[bt] at (axis cs:1,5.7) {Agg};
            \node[font=\footnotesize\rmfamily\bfseries, color=black, anchor=south, yshift=2pt] at (axis cs:1,7.6) {7.6 GB};

            \filldraw[fill=c2D, draw=c2O, line width=0.5pt] (axis cs:1.56,0) rectangle (axis cs:2.44,5.5);   \node[bt] at (axis cs:2,2.75) {Model};
            \filldraw[fill=c2L, draw=c2O, line width=0.5pt] (axis cs:1.56,5.5) rectangle (axis cs:2.44,13.5); \node[bt] at (axis cs:2,9.5) {Distill};
            \node[font=\footnotesize\rmfamily\bfseries, color=black, anchor=south, yshift=2pt] at (axis cs:2,13.5) {13.5 GB};

            \filldraw[fill=c3D, draw=c3O, line width=0.5pt] (axis cs:2.56,0) rectangle (axis cs:3.44,7.5);   \node[bt] at (axis cs:3,3.75) {Repo};
            \filldraw[fill=c3L, draw=c3O, line width=0.5pt] (axis cs:2.56,7.5) rectangle (axis cs:3.44,19.2); \node[bt] at (axis cs:3,13.35) {Align};
            \node[font=\footnotesize\rmfamily\bfseries, color=black, anchor=south, yshift=2pt] at (axis cs:3,19.2) {19.2 GB};

            \filldraw[fill=c4D, draw=c4O, line width=0.5pt] (axis cs:3.56,0) rectangle (axis cs:4.44,6.8);   \node[bt] at (axis cs:4,3.4) {Homoph};
            \filldraw[fill=c4L, draw=c4O, line width=0.5pt] (axis cs:3.56,6.8) rectangle (axis cs:4.44,15.8); \node[bt] at (axis cs:4,11.3) {Feat};
            \node[font=\footnotesize\rmfamily\bfseries, color=black, anchor=south, yshift=2pt] at (axis cs:4,15.8) {15.8 GB};

            \filldraw[fill=c5D, draw=c5O, line width=0.5pt] (axis cs:4.56,0) rectangle (axis cs:5.44,8.5);   \node[bt] at (axis cs:5,4.25) {Hierar};
            \filldraw[fill=c5L, draw=c5O, line width=0.5pt] (axis cs:4.56,8.5) rectangle (axis cs:5.44,18.4); \node[bt] at (axis cs:5,13.45) {VAE};
            \node[font=\footnotesize\rmfamily\bfseries, color=black, anchor=south, yshift=2pt] at (axis cs:5,18.4) {18.4 GB};

            \filldraw[fill=c6D!70!black, draw=c6O, line width=0.5pt] (axis cs:5.56,0) rectangle (axis cs:6.44,3.6);  \node[bt] at (axis cs:6,1.8) {Rout};
            \filldraw[fill=c6D, draw=c6O, line width=0.5pt] (axis cs:5.56,3.6) rectangle (axis cs:6.44,6.5);         \node[bt] at (axis cs:6,5.05) {AGMG};
            \filldraw[fill=c6M, draw=c6O, line width=0.5pt] (axis cs:5.56,6.5) rectangle (axis cs:6.44,9.2);         \node[bt] at (axis cs:6,7.85) {Agg};
            \filldraw[fill=c6L, draw=c6O, line width=0.5pt] (axis cs:5.56,9.2) rectangle (axis cs:6.44,11.7);        \node[bt] at (axis cs:6,10.45) {Base};
            \node[font=\footnotesize\rmfamily\bfseries, color=black, anchor=south, yshift=2pt] at (axis cs:6,11.7) {11.7 GB};

            \end{axis}
        \end{tikzpicture}
        \caption{Memory Footprint.}
        \label{fig:memory} %
    \end{subfigure}
    
    \vspace{0.2cm}
    \caption{System efficiency comparison across graph federated learning baselines. FedMPO achieves high throughput and low memory footprint.}
    \label{fig:efficiency}
\end{figure*}

\section{Theoretical Analysis in Details}
\label{app:theory}

\subsection{Convergence Analysis under Reliability-aware Aggregation}

We provide an appendix-level convergence analysis for \textsc{FedMPO} under modality missingness and reliability-aware aggregation. 
The purpose of this analysis is not to claim a stronger convergence rate than standard federated optimization, but to clarify how modality missingness affects stochastic variance and how reliability-aware aggregation reduces the contribution of unreliable client updates.

\paragraph{Setup.}
Let $F_k(\theta)$ be the local objective of client $k$, and let the standard data-size weight be $p_k=|\mathcal V_k|/\sum_{j=1}^{K}|\mathcal V_j|$. 
The global objective is $F(\theta)=\sum_{k=1}^{K}p_kF_k(\theta)$. 
At communication round $t$, client $k$ performs $E$ local steps and returns the local model $\theta_k^{t+1}$. 
Equivalently, we denote its effective local update direction by
\begin{equation}
\mathbf g_k^t
=
\frac{\theta^t-\theta_k^{t+1}}{\eta_l E},
\end{equation}
where $\eta_l$ is the local learning rate. 
Following the main method, each client also reports lightweight reliability statistics, including average recovery uncertainty $\bar u_k^t$, reconstruction error $\bar e_k^t$, and missing ratio $\rho_k^t$. 
The reliability score and aggregation weight are
\begin{equation}
s_k^t
=
\exp(-\eta_u\bar u_k^t-\eta_e\bar e_k^t-\eta_\rho\rho_k^t),
\quad
\omega_k^t
=
\frac{p_ks_k^t}{\sum_{j=1}^{K}p_js_j^t+\epsilon}.
\end{equation}
The server update can then be written as
\begin{equation}
\theta^{t+1}
=
\sum_{k=1}^{K}\omega_k^t\theta_k^{t+1}
=
\theta^t-\eta_lE\sum_{k=1}^{K}\omega_k^t\mathbf g_k^t .
\end{equation}
For analysis, define the reliability-weighted objective at round $t$ as
\begin{equation}
F_{\omega^t}(\theta)=\sum_{k=1}^{K}\omega_k^tF_k(\theta).
\end{equation}

\paragraph{Assumption D.1 (Smoothness and lower boundedness).}
Each local objective $F_k$ is $L$-smooth, and the global objective is lower bounded by $F^\star$.

\paragraph{Assumption D.2 (Missing-aware stochastic variance).}
The stochastic update direction is an unbiased estimator of the local update direction, and its variance is amplified by modality unreliability:
\begin{equation}
\mathbb E[\mathbf g_k^t]=\bar{\mathbf g}_k^t,
\quad
\mathbb E\left[\|\mathbf g_k^t-\bar{\mathbf g}_k^t\|^2\right]
\leq
\frac{\sigma_k^2(t)}{E},
\end{equation}
where
\begin{equation}
\sigma_k^2(t)
\leq
\sigma^2
\left(
1+\gamma_u\bar u_k^t+\gamma_e\bar e_k^t+\gamma_\rho\rho_k^t
\right).
\end{equation}
This assumption captures the intuition that clients with higher missingness or less reliable recovery produce noisier local updates.

\paragraph{Assumption D.3 (Bounded local drift).}
The bias between the expected local update direction and the local full gradient is bounded by
\begin{equation}
\left\|
\bar{\mathbf g}_k^t-\nabla F_k(\theta^t)
\right\|^2
\leq
C_{\mathrm{dr}}\eta_l^2E^2\delta_k^2,
\end{equation}
where $\delta_k$ measures the client-specific distribution drift and $C_{\mathrm{dr}}$ is a constant.

\paragraph{Theorem D.4 (One-round descent of \textsc{FedMPO}).}
Under Assumptions D.1--D.3, if $\eta_lE\leq 1/L$, the update of \textsc{FedMPO} satisfies
\begin{equation}
\begin{aligned}
\mathbb E
\left[
F_{\omega^t}(\theta^{t+1})
\right]
&\leq
F_{\omega^t}(\theta^t)
-
\frac{\eta_lE}{2}
\left\|
\nabla F_{\omega^t}(\theta^t)
\right\|^2
\\
&\quad
+
\frac{L\eta_l^2E^2}{2}
\sum_{k=1}^{K}
(\omega_k^t)^2
\frac{\sigma_k^2(t)}{E}
+
C_{\mathrm{dr}}L\eta_l^2E^2
\sum_{k=1}^{K}
\omega_k^t\delta_k^2 .
\end{aligned}
\end{equation}
Consequently, summing over $T$ rounds gives
\begin{equation}
\begin{aligned}
\frac{1}{T}
\sum_{t=0}^{T-1}
\mathbb E
\left[
\left\|
\nabla F_{\omega^t}(\theta^t)
\right\|^2
\right]
&\leq
\mathcal O
\left(
\frac{F_{\omega^0}(\theta^0)-F^\star}{\eta_lET}
\right)
\\
&\quad+
\mathcal O
\left(
\frac{L\eta_lE}{T}
\sum_{t=0}^{T-1}
\sum_{k=1}^{K}
(\omega_k^t)^2\sigma_k^2(t)
\right)
\\
&\quad+
\mathcal O
\left(
\frac{L\eta_lE}{T}
\sum_{t=0}^{T-1}
\sum_{k=1}^{K}
\omega_k^t\delta_k^2
\right).
\end{aligned}
\end{equation}

\paragraph{Proof.}
Let
\begin{equation}
\mathbf G^t=\sum_{k=1}^{K}\omega_k^t\mathbf g_k^t
\end{equation}
be the aggregated update direction. 
By $L$-smoothness,
\begin{equation}
F_{\omega^t}(\theta^{t+1})
\leq
F_{\omega^t}(\theta^t)
-
\eta_lE
\left\langle
\nabla F_{\omega^t}(\theta^t),
\mathbf G^t
\right\rangle
+
\frac{L\eta_l^2E^2}{2}
\|\mathbf G^t\|^2 .
\end{equation}
Taking expectation and decomposing $\mathbf G^t$ into its mean and stochastic noise yields
\begin{equation}
\mathbb E
\left[
\|\mathbf G^t-\mathbb E[\mathbf G^t]\|^2
\right]
\leq
\sum_{k=1}^{K}
(\omega_k^t)^2
\mathbb E
\left[
\|\mathbf g_k^t-\bar{\mathbf g}_k^t\|^2
\right]
\leq
\sum_{k=1}^{K}
(\omega_k^t)^2
\frac{\sigma_k^2(t)}{E}.
\end{equation}
Moreover, by Assumption D.3,
\begin{equation}
\left\|
\mathbb E[\mathbf G^t]
-
\nabla F_{\omega^t}(\theta^t)
\right\|^2
\leq
C_{\mathrm{dr}}\eta_l^2E^2
\sum_{k=1}^{K}
\omega_k^t\delta_k^2 .
\end{equation}
Substituting these two bounds into the smoothness inequality and using $\eta_lE\leq 1/L$ gives the one-round descent result. 
Summing the descent inequality from $t=0$ to $T-1$ and rearranging terms yields the stated average stationarity bound.

\paragraph{Remark.}
The bound contains two terms directly related to modality heterogeneity. 
First, the stochastic variance term is scaled by $\sigma_k^2(t)$, which increases with recovery uncertainty, reconstruction error, and missing ratio. 
Second, reliability-aware aggregation multiplies this term by $(\omega_k^t)^2$. 
Since $\omega_k^t$ decreases exponentially with $\bar u_k^t$, $\bar e_k^t$, and $\rho_k^t$, unreliable clients contribute less to the aggregated update variance. 
When all clients have similar reliability, $s_k^t$ becomes approximately constant and the rule reduces to standard data-size-weighted FedAvg.

\subsection{Robustness Analysis of Missing-aware Routing}

We next analyze why the missing-aware routing and structural fallback mechanism can reduce the influence of unreliable recovered modalities. 
The analysis is intentionally stated under simple additive-noise assumptions to clarify the role of uncertainty-aware weighting.

\paragraph{Setup.}
Let $\mathbf z_i^\star$ denote the ideal latent representation of node $v_i$. 
For modality $m$, the processed modality representation is modeled as
\begin{equation}
\mathbf f_i^{(m)}
=
\mathbf z_i^\star+\boldsymbol{\epsilon}_i^{(m)},
\end{equation}
where $\boldsymbol{\epsilon}_i^{(m)}$ is a zero-mean error term with variance
\begin{equation}
\mathbb E
\left[
\|\boldsymbol{\epsilon}_i^{(m)}\|^2
\right]
=
v_i^{(m)}.
\end{equation}
For recovered modalities, the variance $v_i^{(m)}$ increases with the recovery uncertainty $u_i^{(m)}$. 
The structure-only fallback representation is
\begin{equation}
E_{\mathrm{struct}}(\mathbf h_{\mathrm{str},i})
=
\mathbf z_i^\star+\boldsymbol{\epsilon}_{\mathrm{str},i},
\quad
\mathbb E
[
\|\boldsymbol{\epsilon}_{\mathrm{str},i}\|^2
]
=
v_{\mathrm{str},i}.
\end{equation}

\paragraph{Theorem D.5 (Error bound of reliability-weighted fusion).}
Let the fused representation be
\begin{equation}
\mathbf r_i
=
(1-\alpha_{\mathrm{fb},i})
\sum_{m=1}^{M}
a_i^{(m)}\mathbf f_i^{(m)}
+
\alpha_{\mathrm{fb},i}
E_{\mathrm{struct}}(\mathbf h_{\mathrm{str},i}),
\end{equation}
where
\begin{equation}
a_i^{(m)}
=
\frac{\exp(-u_i^{(m)})}
{\sum_{m'=1}^{M}\exp(-u_i^{(m')})+\epsilon}.
\end{equation}
If the modality errors are zero-mean and mutually uncorrelated, then
\begin{equation}
\mathbb E
\left[
\|\mathbf r_i-\mathbf z_i^\star\|^2
\right]
\leq
2(1-\alpha_{\mathrm{fb},i})^2
\sum_{m=1}^{M}
(a_i^{(m)})^2v_i^{(m)}
+
2\alpha_{\mathrm{fb},i}^2v_{\mathrm{str},i}.
\end{equation}
Furthermore, if a recovered modality becomes highly unreliable, i.e., $u_i^{(m)}\rightarrow\infty$, then $a_i^{(m)}\rightarrow 0$, and its contribution to the fusion error vanishes.

\paragraph{Proof.}
Substituting the additive-noise model gives
\begin{equation}
\mathbf r_i-\mathbf z_i^\star
=
(1-\alpha_{\mathrm{fb},i})
\sum_{m=1}^{M}
a_i^{(m)}
\boldsymbol{\epsilon}_i^{(m)}
+
\alpha_{\mathrm{fb},i}
\boldsymbol{\epsilon}_{\mathrm{str},i}.
\end{equation}
Using $\|\mathbf a+\mathbf b\|^2\leq 2\|\mathbf a\|^2+2\|\mathbf b\|^2$, we have
\begin{equation}
\begin{aligned}
\mathbb E
[
\|\mathbf r_i-\mathbf z_i^\star\|^2
]
&\leq
2(1-\alpha_{\mathrm{fb},i})^2
\mathbb E
\left[
\left\|
\sum_{m=1}^{M}
a_i^{(m)}
\boldsymbol{\epsilon}_i^{(m)}
\right\|^2
\right]
+
2\alpha_{\mathrm{fb},i}^2
\mathbb E
[
\|\boldsymbol{\epsilon}_{\mathrm{str},i}\|^2
].
\end{aligned}
\end{equation}
Because modality errors are assumed uncorrelated,
\begin{equation}
\mathbb E
\left[
\left\|
\sum_{m=1}^{M}
a_i^{(m)}
\boldsymbol{\epsilon}_i^{(m)}
\right\|^2
\right]
=
\sum_{m=1}^{M}
(a_i^{(m)})^2v_i^{(m)}.
\end{equation}
This proves the bound. 
Finally, since $a_i^{(m)}$ is a softmax over $-u_i^{(m)}$, $u_i^{(m)}\rightarrow\infty$ implies $a_i^{(m)}\rightarrow 0$. 
Therefore, highly uncertain recovered modalities are automatically suppressed in the fused representation.

\paragraph{Corollary D.6 (Structural fallback under severe missingness).}
If most modalities are missing or highly uncertain, and the fallback gate satisfies $\alpha_{\mathrm{fb},i}\rightarrow 1$, then
\begin{equation}
\limsup
\mathbb E
\left[
\|\mathbf r_i-\mathbf z_i^\star\|^2
\right]
\leq
2v_{\mathrm{str},i}.
\end{equation}
Thus, the structural fallback prevents the representation error from being dominated by unbounded recovery noise.

\paragraph{Remark.}
This result explains the design of the uncertainty estimator and the missing-aware router. 
The uncertainty score controls the modality fusion weight, while the fallback gate allows the model to rely more on structure-only information when recovered modalities are unreliable. 
This does not imply that the structural representation is always superior; rather, it provides a bounded fallback when recovered features become highly uncertain.

\subsection{Complexity and Communication Analysis}

We analyze the computational and communication cost of \textsc{FedMPO}. 
Let $N=|\mathcal V|$ and $|\mathcal E|$ denote the number of nodes and edges on a local client, $M$ be the number of modalities, $d$ be the hidden dimension, and $P=|\Theta|$ be the number of model parameters. 
Let $B$ denote the maximum size of the sampled neighborhood context used by the generator, and define $S=M+B$ as the maximum context-bank size per node. 
In practice, $M$ and $B$ are much smaller than $N$.

\paragraph{Local computation.}
The graph encoder requires sparse message passing and feature projection, whose cost is
\begin{equation}
C_{\mathrm{GNN}}
=
\mathcal O(|\mathcal E|d+Nd^2).
\end{equation}
The topology-aware generator performs attention over a bounded context bank rather than all nodes. 
For each node and modality, the attention cost is $\mathcal O(Sd^2+S^2d)$; therefore,
\begin{equation}
C_{\mathrm{AGMG}}
=
\mathcal O\left(
NM(Sd^2+S^2d)
\right).
\end{equation}
The uncertainty estimator and missing-aware experts operate on hidden representations. 
With a constant number of experts, the routing and fusion cost is
\begin{equation}
C_{\mathrm{route}}
=
\mathcal O(NMd^2).
\end{equation}
Thus, the total local computational complexity per round is
\begin{equation}
C_{\mathrm{FedMPO}}
=
\mathcal O
\left(
|\mathcal E|d
+
Nd^2
+
NM(Sd^2+S^2d)
\right).
\end{equation}
When $M$ and $S$ are bounded constants, the complexity is linear in the local graph size.

\paragraph{Communication cost.}
\textsc{FedMPO} does not transmit raw node features, raw graph structures, recovered modality features, or context-bank representations. 
Each client uploads model parameters together with three scalar reliability statistics: average recovery uncertainty, reconstruction error, and missing ratio. 
Therefore, the per-client communication payload is
\begin{equation}
\mathcal B_{\mathrm{FedMPO}}
=
\mathcal O(P+3)
=
\mathcal O(P).
\end{equation}
The additional reliability statistics introduce only constant-size overhead compared with standard model-parameter transmission.

\paragraph{Comparison with global attention and proxy-based paradigms.}
A centralized global cross-modal attention model over all nodes and modalities has at least
\begin{equation}
\mathcal O(N^2M^2d)
\end{equation}
attention cost, because it constructs pairwise interactions among $NM$ modality tokens. 
Proxy-based federated graph methods may additionally transmit proxy node embeddings, resulting in communication cost
\begin{equation}
\mathcal O(P+|V_{\mathrm{proxy}}|d).
\end{equation}
In contrast, \textsc{FedMPO} keeps graph reasoning, modality generation, and expert routing local, and only communicates model parameters and scalar reliability statistics, , as summarized in Table ~\ref{tab:complexity_comparison}.

\begin{table}[htbp]
\centering
\caption{Complexity comparison of multimodal graph federated learning paradigms.}
\label{tab:complexity_comparison}
\resizebox{\textwidth}{!}{%
\begin{tabular}{lcc}
\toprule
\rowcolor[gray]{0.92}
Architecture & Local Computation (per round) & Communication (per round) \\
\midrule
Centralized global attention 
& $\mathcal O(|\mathcal E|d+N^2M^2d)$ 
& $\mathcal O(NMd_{\mathrm{in}}+|\mathcal E|)$ \\
Proxy-based federated graph learning
& $\mathcal O(|\mathcal E|d+Nd^2)$
& $\mathcal O(P+|V_{\mathrm{proxy}}|d)$ \\
\textsc{FedMPO}
& $\mathcal O(|\mathcal E|d+Nd^2+NM(Sd^2+S^2d))$
& $\mathcal O(P)$ \\
\bottomrule
\end{tabular}%
}
\end{table}

\section{Limitations, Future Work, and Broader Impacts}
\label{app:limitations}

Although \textsc{FedMPO} provides a unified framework for missing-modality robust learning on federated multimodal graphs, it still has several limitations. 
First, the topology-aware cross-modal generation module and the missing-aware MoE routing module introduce additional local computation compared with simple FedAvg-style baselines. 
These modules operate on compact hidden representations rather than raw image or text inputs, and \textsc{FedMPO} does not transmit raw graph structures, node attributes, recovered modality features, or context-bank representations. 
Nevertheless, the extra client-side computation may still be non-negligible for very large graphs or resource-constrained clients.

Second, the missing-aware router relies on uncertainty estimation and load-balancing regularization. 
Although the routing objective mitigates expert collapse in our experiments, extremely heterogeneous clients or severe modality missingness may still lead to biased expert usage or suboptimal routing decisions. 
This limitation suggests that more reliable router calibration, client-adaptive routing, and uncertainty-aware expert selection remain important future directions.

Third, our current missingness simulation mainly considers client-level and node-level modality missingness under controlled missing rates and Dirichlet-based non-IID partitions. 
Real-world missingness can be more complex, including modality-dependent missingness, long-tailed missing patterns, missing-not-at-random cases, and clients with nearly isolated modality spaces. 
Such settings may require more specialized missingness modeling, stronger uncertainty calibration, and better mechanisms for distinguishing recoverable missing signals from inherently unreliable observations.

Fourth, our experiments focus on three downstream tasks: node classification, link prediction, and modality retrieval. 
While these tasks cover both graph-centric and modality-centric evaluation, the current framework has not been fully validated on graph-level prediction, temporal graphs, dynamic user-item graphs, or generation-oriented multimodal graph tasks. 
Extending \textsc{FedMPO} to these scenarios would further test its generality and reveal whether topology-aware completion and reliability-aware aggregation remain effective under more diverse deployment conditions.

Future work can improve \textsc{FedMPO} along several directions. 
One direction is to design more efficient topology-aware generators, such as lightweight sampling-based generators, parameter-efficient adapters, or subgraph-level approximation strategies, to reduce local computation while preserving graph-conditioned modality recovery. 
Another direction is to enhance the missing-aware router with meta-learning, stochastic routing, or calibration-aware objectives, so that it can adapt more robustly to client-specific missingness patterns. 
It would also be useful to study privacy-preserving variants of reliability-aware aggregation, for example by combining scalar reliability statistics with secure aggregation or differentially private noise when client-level missingness information is sensitive. 
Finally, evaluating \textsc{FedMPO} on larger real-world distributed multimodal graph datasets, dynamic graph environments, and application-specific federated platforms would provide stronger evidence for practical deployment.

\textbf{Broader impacts.}
\textsc{FedMPO} aims to improve federated multimodal graph learning under incomplete modalities, which may bring positive impacts to privacy-sensitive and data-fragmented applications. 
For example, it can help different platforms, institutions, or organizations collaboratively train graph learning models without directly sharing raw multimodal data, which is useful for recommendation, information retrieval, social network analysis, and other graph-based services where text, image, and relational data are naturally distributed. 
By explicitly modeling missing modalities and unreliable recovered signals, \textsc{FedMPO} may also improve the robustness of learning systems deployed in environments where data collection is incomplete, uneven, or constrained by privacy and resource limitations.

However, the same capability may also introduce potential risks. 
First, more robust federated multimodal graph learning could be misused in privacy-sensitive profiling, targeted recommendation, or surveillance-like applications if deployed without proper governance. 
Second, although \textsc{FedMPO} does not transmit raw graph structures or raw multimodal attributes, model updates and client-level reliability statistics may still contain indirect information about client data distributions or missingness patterns. 
Third, modality completion may amplify existing biases if missing modalities are not missing at random, especially when some user groups, products, or entities systematically lack certain modalities. 
In such cases, recovered features may create over-confident but inaccurate representations and lead to unfair downstream predictions.

To mitigate these concerns, practical deployments should combine \textsc{FedMPO} with privacy-preserving aggregation, access control, fairness auditing, and uncertainty-aware decision rules. 
When reliability statistics are sensitive, secure aggregation or differential privacy can be used to reduce leakage risks. 
When recovered modalities are used for high-stakes decisions, the system should expose uncertainty estimates, avoid treating generated modalities as ground-truth observations, and include human or domain-expert review when appropriate. 
Overall, \textsc{FedMPO} is intended as a foundational learning framework rather than a decision-making system by itself, and its societal impact depends strongly on the downstream application, deployment constraints, and governance mechanisms.

\section{Hyperparameters in Details}
\label{app:hyperparameters}

In this section, we describe the hyperparameter configurations used in our experiments for \textsc{FedMPO} and comparative baselines. 
To ensure fairness and reproducibility, we distinguish between task-level hyperparameters, which are shared across models within the same downstream task, and model-level hyperparameters, which are specific to individual architectures. 
Unless otherwise specified, all methods use the same modality encoders, client partitions, missingness protocols, and evaluation splits.

\subsection{Global and Model-level Hyperparameters}

To maintain a consistent comparative environment across federated learning scenarios, we standardize the basic training configuration. 
All models are optimized with Adam on local clients. 
Common hyperparameters such as the learning rate, local epochs, client fraction, hidden dimension, and gradient clipping norm are shared across methods whenever applicable. 
The complete set of global hyperparameters is summarized in Table~\ref{tab:global_hyperparameters}.

For conventional federated graph baselines, including FedAvg-Zero, FedGCN, and FedGraphSAGE, we use two graph layers and set the hidden dimensionality to $d=256$. 
For multimodal and missing-modality baselines, we use the same input modality features as \textsc{FedMPO} and tune only model-specific components following the same validation protocol.

\begin{table}[htbp]
\centering
\caption{Global shared hyperparameters.}
\label{tab:global_hyperparameters}
\renewcommand{\arraystretch}{1.1}
\begin{tabular}{l l p{8.5cm}}
\toprule
Hyperparameter & Value & Description \\
\midrule
Optimizer & Adam & Optimizer used for local client training. \\
Learning rate & 0.005 & Initial learning rate for local optimization. \\
Local epochs & 3 & Number of local training epochs per communication round. \\
Client fraction & 1.0 & Fraction of clients sampled per communication round. \\
Hidden dimension & 256 & Shared hidden dimension after modality projection. \\
Gradient clipping norm & 1.0 & Maximum norm for gradient clipping during local training. \\
Warmup rounds & 10 & Number of warmup rounds used for stable early optimization. \\
\bottomrule
\end{tabular}
\end{table}

\subsection{\textsc{FedMPO} Specific Hyperparameters}

\textsc{FedMPO} contains three main components: topology-aware cross-modal generation, missing-aware multimodal routing, and reliability-aware federated aggregation. 
The corresponding hyperparameters are summarized in Table~\ref{tab:fedmpo_specific_hyperparams}. 
The reconstruction coefficient $\lambda_{\mathrm{rec}}$, alignment coefficient $\lambda_{\mathrm{align}}$, and routing coefficient $\lambda_{\mathrm{route}}$ are tuned on validation splits and fixed for all test runs under the same task setting.

\begin{table}[htbp]
\centering
\caption{\textsc{FedMPO} specific hyperparameters.}
\label{tab:fedmpo_specific_hyperparams}
\small
\renewcommand{\arraystretch}{1.1}
\begin{tabularx}{\textwidth}{@{} l l X @{}}
\toprule
Hyperparameter & Value & Description \\
\midrule
\rowcolor[gray]{0.9} \multicolumn{3}{c}{\textbf{Topology-aware Cross-modal Generation}} \\
Zero impute full & True / 1 & Whether to use zero imputation when all modalities of a node are unavailable. \\
Use missing tokens & False & Whether to use learnable missing-modality tokens. \\
Use impute then pass & True & Whether missing features are imputed before graph context encoding. \\
Missing skip $\alpha$ & 0.35 & Residual mixing ratio between imputed and aggregated features. \\
Number of attention heads & 4 & Number of heads for cross-modal multi-head attention. \\
Attention warmup rounds & 30 & Number of warmup rounds for gradually enabling attention-based generation. \\
Interpolation $\alpha$ bounds & $[0.55, 0.78]$ & Dynamic coefficient bounds for combining generated features with structural priors. \\

\rowcolor[gray]{0.9} \multicolumn{3}{c}{\textbf{Missing-aware Routing and Fusion}} \\
Use conditional routing & True & Whether to enable conditional routing in missing-aware MoE fusion. \\
Router hidden dimension & $\max(4,d_{\mathrm{model}}/2)$ & Hidden dimension for the router and uncertainty estimator. \\
Uncertainty clamp max & 0.12 & Maximum clamping value for uncertainty-based penalty. \\
Router temperature $\tau$ & 1.0 & Softmax temperature for router probabilities. \\
Uniform floor & 0.0 & Minimum probability floor for routing experts. \\
Dropout & 0.1 & Dropout rate before the final task prediction layer. \\

\rowcolor[gray]{0.9} \multicolumn{3}{c}{\textbf{Reliability-aware Aggregation}} \\
Uncertainty coefficient $\eta_u$ & 1.0 & Weight of average recovery uncertainty in client reliability scoring. \\
Reconstruction coefficient $\eta_e$ & 1.0 & Weight of reconstruction error in client reliability scoring. \\
Missing-ratio coefficient $\eta_\rho$ & 1.0 & Weight of client-level missing ratio in client reliability scoring. \\
\bottomrule
\end{tabularx}
\end{table}

\subsection{Baseline Specific Configurations}

For baseline methods, we follow their standard architectural choices whenever possible and use the same hidden dimension, modality encoders, client splits, and missingness protocol as \textsc{FedMPO}. 
Specific hyperparameter settings for representative baselines are summarized in Table~\ref{tab:baselines_hyperparameters}. 
For baselines that are not naturally designed for all downstream tasks, we keep their original learning objectives and report N/A when the original method is not directly applicable without changing its core formulation.

\begin{table}[htbp]
\centering
\caption{Baseline specific hyperparameters.}
\label{tab:baselines_hyperparameters}
\small 
\renewcommand{\arraystretch}{1.1}
\begin{tabularx}{\textwidth}{@{} l l X @{}}
\toprule
Hyperparameter & Value & Description \\
\midrule
\rowcolor[gray]{0.9} \multicolumn{3}{c}{\textbf{FedGCN}} \\
Number of layers & 2 & Number of GCNConv layers. \\
Dropout & 0.5 & Dropout rate applied during local training. \\

\rowcolor[gray]{0.9} \multicolumn{3}{c}{\textbf{FedGraphSAGE}} \\
Number of layers & 2 & Number of GraphSAGE layers. \\
Dropout & 0.3 & Dropout rate for the local GraphSAGE encoder. \\

\rowcolor[gray]{0.9} \multicolumn{3}{c}{\textbf{FedPub}} \\
$\ell_1$ coefficient & $1\times 10^{-3}$ & $L_1$ regularization coefficient for the personalized subgraph mask. \\
Dropout & 0.5 & Dropout rate applied after MaskedGCN layers. \\
Mask initialization & Normal & Normal distribution initialization for the personalized mask. \\

\rowcolor[gray]{0.9} \multicolumn{3}{c}{\textbf{FedMVP}} \\
Number of attention heads & 4 & Number of heads for cross-modal multi-head attention. \\
Dropout & 0.1 & Dropout rate applied during attention and local training. \\

\rowcolor[gray]{0.9} \multicolumn{3}{c}{\textbf{FedMAC}} \\
Number of attention heads & 4 & Number of heads for cross-modal multi-head attention. \\
Dropout & 0.1 & Dropout rate applied during cross-modal aggregation. \\

\rowcolor[gray]{0.9} \multicolumn{3}{c}{\textbf{FedProto}} \\
Initial logit scale & 10.0 & Initial scaling factor for prototype matching logits. \\
Logit scale bounds & $[1.0,20.0]$ & Clamping bounds for the logit scaling factor. \\

\rowcolor[gray]{0.9} \multicolumn{3}{c}{\textbf{FedLAP}} \\
Refinement mix ratio & 0.7 / 0.3 & Mix ratio between original features and Laplacian-refined features. \\

\rowcolor[gray]{0.9} \multicolumn{3}{c}{\textbf{S2FGL, FedSPA, and FedIIH}} \\
Default configuration & Official setting & We follow the reported configurations and tune shared training parameters using the same validation protocol. \\
\bottomrule
\end{tabularx}
\end{table}

\subsection{Task-level Hyperparameters}

Experiments are conducted across multiple communication rounds with local training iterations to handle data heterogeneity. 
The task-level hyperparameters are tailored to the requirements of node classification, link prediction, and modality retrieval. 
The detailed configurations, including loss functions, loss weights, and evaluation metrics, are summarized in Table~\ref{tab:task_hyperparameters}.

For all tasks, raw image or video features are encoded into 512-dimensional embeddings, while textual features are encoded into 768-dimensional embeddings. 
These modality features are then projected into the shared hidden space with dimension $d=256$.

\begin{table}[htbp]
\centering
\caption{Task specific hyperparameters.}
\label{tab:task_hyperparameters}
\small 
\renewcommand{\arraystretch}{1.1}
\begin{tabularx}{\textwidth}{@{} l l X @{}}
\toprule
Hyperparameter & Value & Description \\
\midrule
\rowcolor[gray]{0.9} \multicolumn{3}{c}{\textbf{Node Classification (NC)}} \\
Task loss function & Cross Entropy & Default cross-entropy loss used for node classification. \\
Evaluation metrics & Accuracy / F1-score & Metrics used for node classification. \\
$\lambda_{\mathrm{rec}}$ & 0.05 & Weight for the reconstruction loss. \\
$\lambda_{\mathrm{align}}$ & 0.01 & Weight for the cross-modal alignment loss. \\
$\lambda_{\mathrm{route}}$ & 0.01 & Weight for uncertainty calibration and expert load balancing. \\

\rowcolor[gray]{0.9} \multicolumn{3}{c}{\textbf{Link Prediction (LP)}} \\
Global rounds & 150 & Total number of federated communication rounds. \\
Task loss function & BCE + BPR + Margin & Combination of binary classification and ranking objectives. \\
Evaluation metrics & AUC / AP & Metrics used for link prediction. \\
Loss weight (BCE) & 1.0 & Weight for Binary Cross-Entropy loss. \\
Loss weight (BPR) & 0.5 & Weight for Bayesian Personalized Ranking loss. \\
Loss weight (Margin) & 0.3 & Weight for Margin Ranking loss. \\
Margin value & 0.1 & Margin size for the Margin Ranking loss. \\
Hard negative pool scale & 4.0 & Multiplier for scaling the hard negative sampling pool size. \\
Hard negative min pool & 256 & Minimum number of samples in the hard negative pool. \\
$\lambda_{\mathrm{rec}}$ & 0.05 & Weight for the reconstruction loss. \\
$\lambda_{\mathrm{align}}$ & 0.01 & Weight for the cross-modal alignment loss. \\
$\lambda_{\mathrm{route}}$ & 0.01 & Weight for uncertainty calibration and expert load balancing. \\

\rowcolor[gray]{0.9} \multicolumn{3}{c}{\textbf{Modality Retrieval (MR)}} \\
Query modality & \texttt{image} & Query modality for retrieval. \\
Gallery modality & \texttt{text} & Target/gallery modality for retrieval. \\
Task loss function & InfoNCE & Contrastive loss used for cross-modal retrieval. \\
Evaluation metrics & Recall@5 / MRR & Metrics used for modality retrieval. \\
InfoNCE temperature $\tau$ & 0.07 & Temperature scaling parameter for InfoNCE. \\
$\lambda_{\mathrm{rec}}$ & 0.5 & Weight for the reconstruction loss. \\
$\lambda_{\mathrm{align}}$ & 0.01 & Weight for the cross-modal alignment loss. \\
$\lambda_{\mathrm{route}}$ & 0.01 & Weight for uncertainty calibration and expert load balancing. \\
Auxiliary classification weight $\lambda_{\mathrm{cls}}$ & 0.2 & Weight for the auxiliary classification objective when used. \\
\bottomrule
\end{tabularx}
\end{table}
\end{document}